%% file: arxiv.tex
\documentclass[10pt,twocolumn,letterpaper]{article}

\usepackage[pagenumbers]{cvpr} %

\input{preamble}

\definecolor{cvprblue}{rgb}{0.21,0.49,0.74}
\definecolor{RowColor}{rgb}{0.93, 0.98, 0.97}
\definecolor{deepgray}{rgb}{0.5, 0.5, 0.5}
\usepackage[utf8]{inputenc} %
\usepackage[T1]{fontenc}    %
\usepackage{url}            %
\usepackage{booktabs}       %
\usepackage{hhline}
\usepackage{amsbsy,amsmath}
\usepackage{amsfonts}       %
\usepackage{nicefrac}       %
\usepackage{microtype}      %
\usepackage{colortbl}
\usepackage{xcolor}         %
\usepackage{graphicx}
\usepackage{amsmath}
\usepackage{amssymb}
\usepackage{marvosym}
\usepackage{booktabs}
\usepackage{multirow}
\usepackage{makecell}
\usepackage{url}            %
\usepackage{booktabs}       %
\usepackage{amsfonts}       %
\usepackage{nicefrac}       %
\usepackage{microtype}      %
\usepackage{xcolor}         %
\usepackage{enumitem}
\usepackage{graphicx}
\usepackage{multirow}
\usepackage{amsmath}
\usepackage{amstext}
\usepackage{MnSymbol}
\usepackage{wasysym}
\usepackage{listings}
\usepackage{color,colortbl}
\usepackage{wrapfig}
\usepackage{soul}
\usepackage{arydshln}
\usepackage{bm}
\usepackage{algorithm}
\usepackage{algpseudocode}
\usepackage{textcomp}
\usepackage{multirow}
\usepackage{wrapfig}
\usepackage{subcaption}
\usepackage{tabularx}
\usepackage{adjustbox}
\usepackage{makecell}
\usepackage{array}
\usepackage{fontawesome5}
\usepackage{amssymb}%
\usepackage{pifont}%
\usepackage{graphicx}
\usepackage[pagebackref,breaklinks,colorlinks,citecolor=cvprblue]{hyperref}

\definecolor{LightBlue}{rgb}{1,0.9,0.9}
\definecolor{LightGreen}{rgb}{0.9,1,0.9}
\definecolor{lightRed}{rgb}{0.9,0.9,1}
\definecolor{LightPurple}{rgb}{0.9,0.8,1}
\definecolor{SoftCoral}{rgb}{0.94, 0.8, 0.7}
\definecolor{MintCream}{rgb}{0.7, 0.9, 0.95}
\definecolor{Honeydew}{rgb}{0.6, 0.9, 0.94}
\definecolor{AzureMist}{rgb}{0.7, 0.7, 0.7}
\definecolor{LavenderBlush}{rgb}{1, 0.94, 0.96}
\definecolor{codedark}{rgb}{1,0.7,0.8}

\def\paperID{6878} %
\def\confName{CVPR}
\def\confYear{2024}

\title{Rewrite the Stars\raisebox{-0.3em}{\includegraphics[height=1.2em]{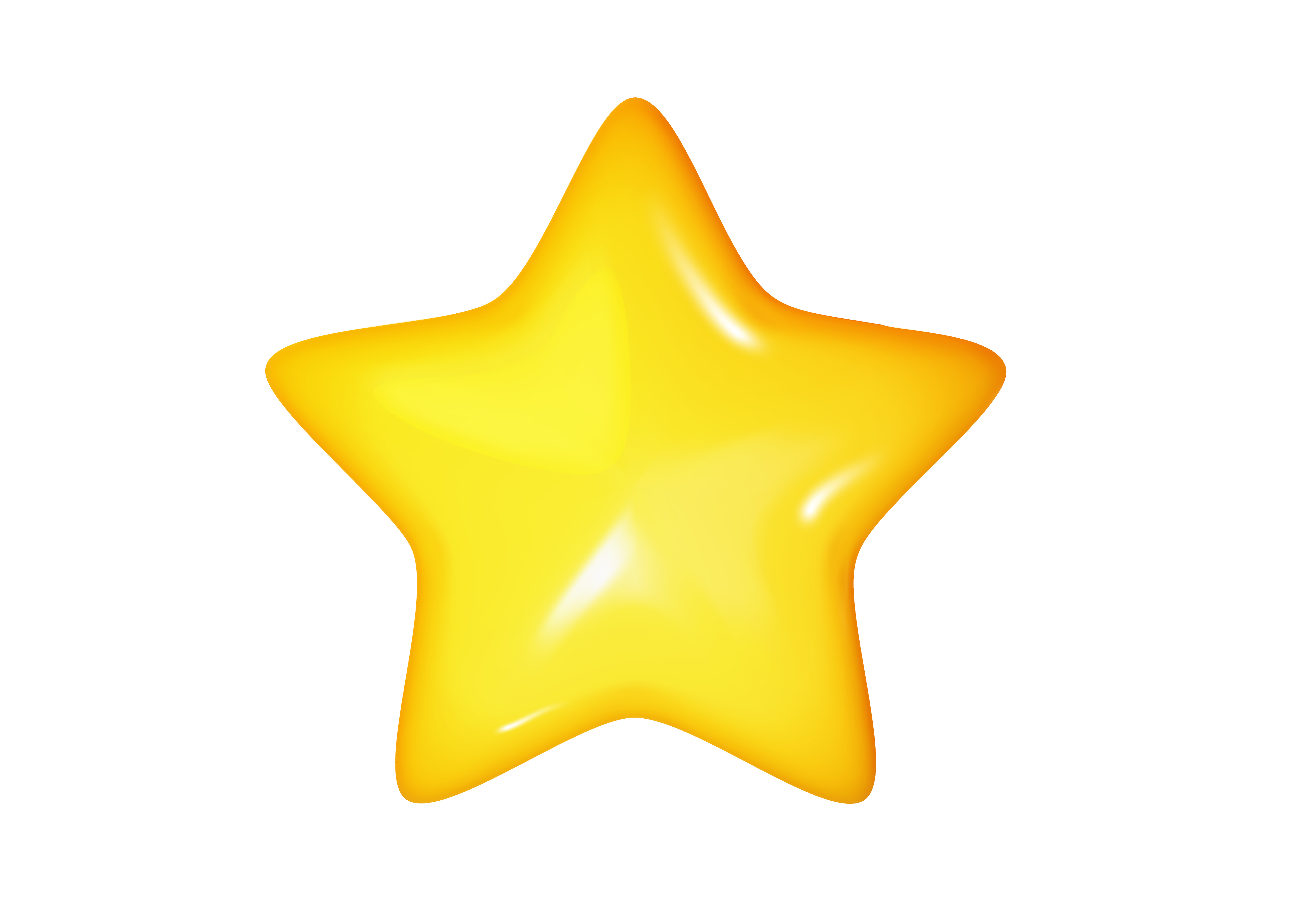}}}

\author{
Xu Ma\textsuperscript{1}, Xiyang Dai\textsuperscript{2}, Yue Bai\textsuperscript{1},
Yizhou Wang\textsuperscript{1}, Yun Fu\textsuperscript{1}
\\
{ \textsuperscript{1}Northeastern University  \ \ \ \ \ \ \ \textsuperscript{2}Microsoft
}\\
}

\begin{document}
\maketitle
\begin{abstract}

Recent studies have drawn attention to the untapped potential of the "star operation" (element-wise multiplication) in network design. While intuitive explanations abound, the foundational rationale behind its application remains largely unexplored. Our study attempts to reveal the star operation's ability to map inputs into high-dimensional, non-linear feature spaces—akin to kernel tricks—without widening the network. We further introduce StarNet, a simple yet powerful prototype, demonstrating impressive performance and low latency under compact network structure and efficient budget. Like stars in the sky, the star operation appears unremarkable but holds a vast universe of potential. Our work encourages further exploration across tasks, with codes available at \href{https://github.com/ma-xu/Rewrite-the-Stars}{https://github.com/ma-xu/Rewrite-the-Stars}.

\end{abstract}

\section{Introduction}

The learning paradigms have imperceptibly and gradually evolved in the past decade. Since AlexNet~\cite{krizhevsky2012imagenet}, a myriad of deep networks~\cite{simonyan2014very,he2016deep,huang2017densely,brock2021high,liu2022convnet} have emerged, each building on the other. Despite their characteristic insights and contributions, this line of models is mostly based on the blocks that blend linear projection (\textit{i.e.}, convolution and linear layers) with non-linear activations. Since~\cite{vaswani2017attention}, self-attention has dominated natural language processing, and later computer vision~\cite{dosovitskiy2020image}. The most distinctive feature of self-attention is mapping features to different spaces and then constructing an attention matrix through dot-product multiplication. However, this implementation is not efficient, and results in the attention complexity scaling quadratically with the increase in the number of tokens.

\begin{figure}
    \centering
    \includegraphics[width=0.998\linewidth]{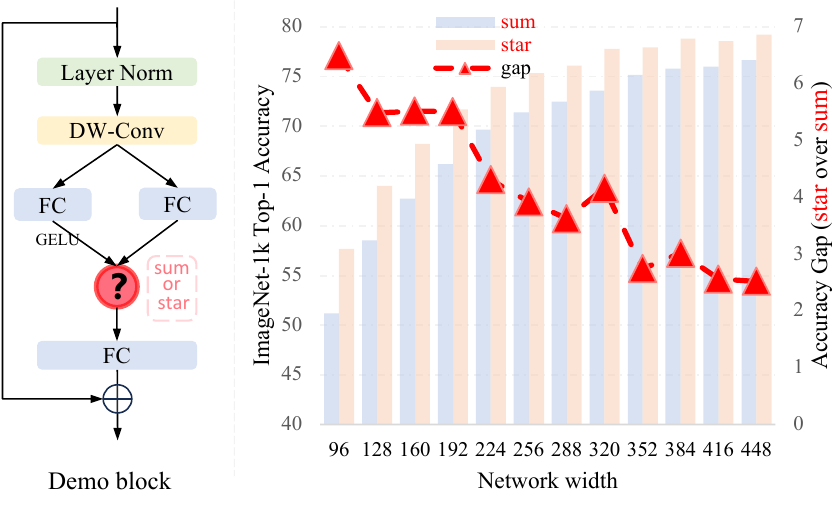}
    \caption{\textbf{Illustration of the advantage of the star operation} (element-wise multiplication). The left side depicts a basic building block abstracted from related works~\cite{yang2022focal,guo2023visual,rao2022hornet}, with ``\textcolor{red}{{\sffamily ?}}" representing either `star' or `summation.' The right side highlights the notable performance disparity between the two operations, with `star' exhibiting superior performance, particularly with a narrower width. Please check Sec.~\ref{sec:Empirical_superiority_of_star_operation} for more results.
    }
    \vspace{-6mm}
    \label{fig:teaser}
\end{figure}

Recently, a new learning paradigm has been gaining increased attention: fusing different subspace features through element-wise multiplication. We refer to this paradigm as `star operation' for simplicity (owing to the element-wise multiplication symbol resembling a star). Star operation exhibits promising performance and efficiency across various research fields, including Natural Language Processing (\textit{i.e.}, Monarch Mixer~\cite{fu2023monarch}, Mamba~\cite{gu2023mamba}, Hyena Hierarchy~\cite{poli2023hyena}, GLU~\cite{shazeer2020glu}, \textit{etc.}), Computer Vision (\textit{i.e.}, FocalNet~\cite{yang2022focal}, HorNet~\cite{rao2022hornet}, VAN~\cite{guo2023visual}, \textit{etc.}), and more. For illustration, we construct a `demo block' for image classification, as depicted on the left in Fig.~\ref{fig:teaser}. By stacking multiple demo blocks following a stem layer, we construct a straightforward model named DemoNet.  Keeping all other factors constant, we observe that element-wise multiplication (star operation) consistently surpasses summation in performance, as illustrated on the right in Fig.~\ref{fig:teaser}.
Although the star operation is remarkably simple, it raises the question: why does it deliver such gratifying results? In response to this, several assumptive explanations have been proposed.
For instance, FocalNet~\cite{yang2022focal} posits that star operations can act as modulation or gating mechanisms, dynamically altering input features. HorNet~\cite{rao2022hornet} suggests that the advantage lies in harnessing high-ordered features. Meanwhile, both VAN~\cite{guo2023visual} and Monarch Mixer~\cite{fu2023monarch} attribute the effectiveness to convolutional attention.
While these preliminary explanations provide some insights, they are largely based on intuition and assumptions, lacking comprehensive analysis and strong evidence. Consequently, the foundational rationale behind remain unexamined, posing a challenge to better understanding and effectively leveraging the star operations.

In this work, we explain the strong representative ability of star operation by explicitly demonstrating that: the star operation possesses the capability to map inputs into an exceedingly high-dimensional, non-linear feature space. 
Instead of relying on intuitive or assumptive high-level explanations, we delve deep into the details of star operations.
By rewriting and reformulating star operations, we uncover that this seemingly simple operation can generate a new feature space comprising approximately  $\left(\frac{d}{\sqrt{2}}\right)^2$ linearly independent dimensions, as detailed in Sec.~\ref{sec:star_operation_in_one_layer}.
The way star operation achieves such non-linear high dimensions is distinct from traditional neural networks that increase the network width (\textit{aka} channel number). 
Rather, the star operation is analogous to kernel functions that conduct pairwise multiplication of features across distinct channels, particularly polynomial kernel functions~\cite{shawe2004kernel,hofmann2008kernel}.
When incorporated into neural networks and with the stacking of multiple layers, each layer contributes to an exponential increase in the implicit dimensional complexity. 
With just a few layers, the star operation enables the attainment of nearly infinite dimensions within a compact feature space, as elaborated in Sec~.\ref{sec:generalized_to_multiple_layers}.
Operating within a compact feature space while benefiting from implicit high dimensions, that is where the star operation captivates with its unique charm.

Drawing from the aforementioned insights, we infer that the star operation may inherently be more suited to efficient, compact networks as opposed to the conventionally used large models.
To validate this, we introduce a proof-of-concept efficient network, StarNet, characterized by its conciseness and efficiency. The detailed architecture of StarNet can be found in Fig.~\ref{fig:starnet_architecture}.
StarNet is notably straightforward, devoid of sophisticated designs and fine-tuned hyper-parameters. 
In terms of design philosophy, StarNet clearly diverges from existing networks, as illustrated in Table~\ref{tab:summary_efficient_networks}.
Capitalizing on the efficacy of the star operation, our StarNet can even surpass various meticulously designed efficient models, like MobileNetv3~\cite{howard2019searching}, EdgeViT~\cite{pan2022edgevits}, FasterNet~\cite{chen2023run}, \textit{etc}. 
For example, our StarNet-S4 outperforms EdgeViT-XS by 0.9\% top-1 accuracy on ImageNet-1K validation set while running $3\times$ faster on iPhone13 and CPU, and $2\times$ faster on GPU.
These results not only empirically validate our insights regarding the star operation but also underscore its practical value in real-world applications.

We succinctly summarize and emphasize the key contributions of this work as follows:
\begin{itemize}
    \item Foremost, we demonstrated the effectiveness of star operations, as illustrated in Fig.~\ref{fig:teaser}. We unveiled that the star operation possesses the capability to project features into an exceedingly high-dimensional implicit feature space, akin to polynomial kernel functions, as detailed in Sec.~\ref{sec:rewrite_the_stars}.
    \item We validated our analysis through empirical results (refer Fig.~\ref{fig:teaser}, Table~\ref{tab:demonet_width}, and Table~\ref{tab:demonet_depth}, \textit{etc.}), theoretical exploration (in Sec.~\ref{sec:rewrite_the_stars}), and visual representation (see Fig.~\ref{fig:2d_points}). 
    \item Drawing inspiration from our analysis, we identify the utility of the star operation in the realm of efficient networks and present a proof-of-concept model, StarNet. Remarkably, StarNet achieves promising performance without the need for intricate designs or meticulously selected hyper-parameters, surpassing numerous efficient designs.
    \item It is noteworthy that there exists a multitude of unexplored possibilities based on the star operation, such as learning without activations and refining operations within implicit dimensions. We envision that our analysis can serve as a guiding framework, steering researchers away from haphazard network design attempts.
\end{itemize}

\begin{table}[]
    \centering
    {\small
    \begin{tabular}{l@{\extracolsep{4pt}}!{\vrule width 1.0pt}l}
        \textbf{Main Insight} &  \multicolumn{1}{c}{\textbf{Networks}}\\
        \Xhline{3\arrayrulewidth}
        DW-Conv & \textcolor{deepgray}{ MobileNetv2~\cite{howard2017mobilenets,sandler2018mobilenetv2}}\\
        Feature Shuffle & \textcolor{deepgray}{ShuffleNet~\cite{zhang2018shufflenet}, ShuffleNetv2~\cite{ma2018shufflenet} }\\
        Feature Re-use & \textcolor{deepgray}{GhostNet~\cite{han2020ghostnet}, FasterNet~\cite{chen2023run} }\\
        NAS  & \textcolor{deepgray}{EfficientNet~\cite{tan2019efficientnet,tan2021efficientnetv2}, MnasNet~\cite{tan2019mnasnet} }\\
        Re-parameterization  & \textcolor{deepgray}{MobileOne~\cite{vasu2023mobileone}, FasterViT~\cite{hatamizadeh2023fastervit} }\\
        Hybrid Architecture &\textcolor{deepgray}{ Mobile-Former~\cite{chen2022mobile}, EdgeViT~\cite{pan2022edgevits}}\\
        \Xhline{3\arrayrulewidth}
        \rowcolor{RowColor}Implicit high dimension  & \multicolumn{1}{c}{StarNet$_{\mathrm{(ours)}}$}
    \end{tabular}
    }
    \vspace{-2mm}
    \caption{\textbf{Taxonomy of prominent efficient networks based on their key insights}. We introduce StarNet that is distinguished by exploring a novel approach: leveraging implicit high dimensions through star operations to enhance network efficiency.}
    \label{tab:summary_efficient_networks}
\end{table}

\section{Related Work}

\paragraph{Element-wise Multiplication in Neural Networks.}
Recent efforts have demonstrated that utilizing element-wise multiplication can be a more effective choice than summation in network design for feature aggregation, as exemplified by FocalNet~\cite{yang2022focal}, VAN~\cite{guo2023visual}, Conv2Former~\cite{hou2022conv2former}, HorNet~\cite{rao2022hornet}, and more~\cite{lin2023scale,li2022efficient,wasim2023video,yun2023spanet}.
To elucidate its superiority, intuitive explanations have been developed, including modulation mechanism, high-order features, and the integration of convolutional attention, \textit{etc}.
Although many \textit{tentative} explanations have been proposed and \textit{empirical} improvements have been achieved, the \textbf{foundational} rationale behind has remained unexamined.  
In this work, we explicitly emphasize that the element-wise multiplication is crucial, regardless of trivial architectural modifications.
It has the capacity to implicitly transform input features into exceptionally high and nonlinear dimensions in a novel manner, but operate in low-dimensional space.

\paragraph{High-Dimensional \& Non-Linear Feature Transformation.}
The inclusion of high-dimensional and nonlinear features is crucial in both traditional machine learning algorithms~\cite{bishop2006pattern,hastie2009elements} and deep learning networks~\cite{lecun2015deep,krizhevsky2012imagenet,zeiler2014visualizing,simonyan2014very}.
This necessity stems from the intricate nature of real-world data and the inherent capacity of models to represent this complexity.
Nevertheless, it is important to recognize that these two lines of approaches achieve this goal from different perspectives. 
In the era of deep learning, we typically start by linearly projecting low-dimensional features into a high-dimensional space and then introduce non-linearity using activation functions (\textit{e.g.}, ReLU, GELU, \textit{etc}.).
In contrast, we can simultaneously attain high-dimensionality and non-linearity using kernel tricks~\cite{shawe2004kernel,cortes1995support} in traditional machine learning algorithms. 
For instance, a polynomial kernel function $k\left(x_1, x_2\right) = \left(\gamma x_1\cdot x_2 + c\right)^d$ can project the input feature $x_1,x_2\in \mathbb{R}^n$ into a $\left(n+1\right)^d$ high-dimensional non-linear feature space; a Gaussian kernel function $k\left(x_1, x_2\right) =  \mathrm{exp}\left(-\left\| x_1\right\|^2\right)\mathrm{exp}\left(-\left\| x_2\right\|^2\right)\sum_{i=0}^{+\infty }\frac{\left ( 2x_1^\intercal x_2 \right )^i}{i!}$ can result in an infinite-dimensional feature space through Taylor expansion.
As a comparison, we can observe that classical machine learning kernel methods and neural networks differ in their implementation and comprehension of high-dimensional and non-linear features. 
In this work, we demonstrate that the star operation can obtain a high-dimensional and non-linear feature space within a low-dimensional input, akin to the principles of kernel tricks. A simple visualization experiment shown in Fig~\ref{fig:2d_points} further illustrates the connections between star operation and polynomial kernel functions.

\paragraph{Efficient Networks.} 
Efficient Networks strive to strike the ideal balance between computational complexity and performance. In recent years, numerous innovative concepts have been introduced to enhance the efficiency of networks. These include depth-wise convolution~\cite{howard2017mobilenets,sandler2018mobilenetv2}, feature re-use~\cite{han2020ghostnet,chen2023run}, and re-parameterization~\cite{vasu2023mobileone}, among others. A comprehensive summary can be found in Table~\ref{tab:summary_efficient_networks}.
In stark contrast to all previous methods, we demonstrate that the star operation can serve as a novel methodology for efficient networks. It has the unique capability to implicitly consider extremely high-dimensional features while performing computations in a low-dimensional space.
The salient merit is discriminative in distinguishing star operation from other technologies in the realm of efficient networks, and makes it particularly suitable for efficient network design. 
With star operations, we demonstrate that a straightforward network can easily outperform heavily handcrafted designs.

\section{Rewrite the Stars}
\label{sec:rewrite_the_stars}
We start by rewriting the star operation to explicitly showcase its ability to achieve exceedingly high dimensions.
We then demonstrate that after multiple layers, star can significantly increase the implicit dimensions to nearly infinite dimensionality.
Discussions are presented subsequently.

\subsection{Star Operation in One layer}
\label{sec:star_operation_in_one_layer}

In a single layer of neural networks, the star operation is typically written as $\left(\mathrm{W}_1^\mathrm{T}\mathrm{X}+\mathrm{B}_1\right) * \left(\mathrm{W}_2^\mathrm{T}\mathrm{X}+\mathrm{B}_2\right)$, signifying the fusion of two linearly transformed features through element-wise multiplication. 
For convenience, we consolidate the weight matrix and bias into a single entity, denoted by  $\mathrm{W} =  \begin{bmatrix} \mathrm{W}\\ \mathrm{B} \end{bmatrix}$, and similarly, $\mathrm{X} =  \begin{bmatrix} \mathrm{X}\\ 1 \end{bmatrix}$, resulting star operation $\left(\mathrm{W}_1^\mathrm{T}\mathrm{X}\right) * \left(\mathrm{W}_2^\mathrm{T}\mathrm{X}\right)$.
To simplify our analysis, we focus on the scenario involving a one-output channel transformation and a single-element input. Specifically, we define $w_1, w_2, x \in \mathbb{R}^{\left(d+1\right)\times 1}$, where $d$ is the input channel number.
It can be readily extended to accommodate multiple output channels $\mathrm{W}_1, \mathrm{W}_2\in\mathbb{R}^{\left(d+1\right)\times \left(d'+1\right)}$ and to handle multiple feature elements, with $\mathrm{X} \in \mathbb{R}^{\left(d+1\right)\times n}$.

Generally, we can rewrite the star operation by:
{\small
\begin{align}
\label{eq:star}
&\: \: \: \: \:\: w_1^\mathrm{T}x * w_2^\mathrm{T}x\\
\label{eq:star_second}
&= \left ( \sum_{i=1}^{d+1}w_1^ix^i \right )*   \left ( \sum_{j=1}^{d+1}w_2^jx^j \right ) \\\
\label{eq:star_sum}
& = \sum_{i=1}^{d+1}\sum_{j=1}^{d+1}w_1^iw_2^jx^ix^j \\
 \label{eq:star_unfold}
 & = \begin{matrix}
\underbrace{\alpha_{\left(1,1\right)}x^1x^1 + \cdots + \alpha_{\left(4,5\right)}x^4x^5 + \cdots + \alpha_{\left(d+1,d+1\right)}x^{d+1}x^{d+1}} \\
{\scriptstyle {\color{gray} \left(d+2\right)\left(d+1\right)/2 \: \: \:\mathrm{items}}}
\end{matrix}
\end{align}
}
where we use $i, j$ to index the channel and $\alpha$ is a coefficient for each item:
\begin{equation}
    \alpha_{\left(i,j\right)} = \left\{\begin{matrix}
 w_1^iw_2^j& \mathrm{if} \: i==j,  \\
 w_1^iw_2^j+ w_1^jw_2^i&  \mathrm{if} \: i!=j.\\
\end{matrix}\right.
\end{equation}

Upon rewriting the star operation delineated in Eq.~\ref{eq:star}, we can expand it into a composition of $\frac{\left(d+2\right)\left(d+1\right)}{2}$ distinct items, as presented in Eq.~\ref{eq:star_unfold}. 
Of note is that each item (besides $\alpha_{\left(d+1,:\right)}x^{d+1}x$) exhibits a non-linear association with $x$, indicating that they are individual and implicit dimensions. 
Therefore, we perform computations within a $d$-dimensional space using a computationally efficient star operation, yet we achieve a representation in a $\frac{\left(d+2\right)\left(d+1\right)}{2} \approx \left(\frac{d}{\sqrt{2}}\right)^2$ (considering $d\gg2$) implicit dimensional feature space, significantly amplifying the feature dimensions without incurring any additional computational overhead within a single layer.
Of note is that this prominent property shares a similar philosophy as kernel functions, and we refer readers to~\cite{hofmann2008kernel,shawe2004kernel} for a wider and deeper outlook.

\subsection{Generalized to multiple layers}
\label{sec:generalized_to_multiple_layers}
Next, we demonstrate that by stacking \textbf{multiple layers}, we can exponentially increase the implicit dimensions to nearly infinite in a recursive manner. 

Considering an initial network layer with a width of $d$, the application of one star operation yields the expression $\sum_{i=1}^{d+1}\sum_{j=1}^{d+1}w_1^iw_2^jx^ix^j$, as detailed in Eq.~\ref{eq:star_sum}. This results in a representation within an implicit feature space of $\mathbb{R}^{\left(\frac{d}{\sqrt{2}}\right)^{2^1}}$.
Let $O_l$ denote the output of $l$-th star operation, we get:
\begin{align}
 O_1& = \sum_{i=1}^{d+1}\sum_{j=1}^{d+1}w_{(1,1)}^iw_{(1,2)}^jx^ix^j \; \; \; \;\; \;\; \; \in \mathbb{R}^{\left(\frac{d}{\sqrt{2}}\right)^{2^1}}\\
O_2& = \mathrm{W}_{2,1}^\mathrm{T}\mathrm{O_1} * \mathrm{W}_{2,2}^\mathrm{T}\mathrm{O_1} \; \; \;\;\; \;\; \; \;\; \;\; \;\; \;\; \; \in \mathbb{R}^{\left(\frac{d}{\sqrt{2}}\right)^{2^2}} \\
O_3& = \mathrm{W}_{3,1}^\mathrm{T}\mathrm{O_2} * \mathrm{W}_{3,2}^\mathrm{T}\mathrm{O_2} \; \;\; \;\; \; \;\; \;\; \;\; \;\; \;\; \; \in \mathbb{R}^{\left(\frac{d}{\sqrt{2}}\right)^{2^3}} \\
\cdots \\
O_l& = \mathrm{W}_{l,1}^\mathrm{T}\mathrm{O_{l-1}} * \mathrm{W}_{l,2}^\mathrm{T}\mathrm{O_{l-1}} \; \; \;\; \;\; \;\; \;\; \;\; \; \in \mathbb{R}^{\left(\frac{d}{\sqrt{2}}\right)^{2^l}} 
\end{align}
That is, with $l$ layers, we can implicitly obtain a feature space belonging to $\mathbb{R}^{\left(\frac{d}{\sqrt{2}}\right)^{2^l}}$. 
For instance, given a 10-layer isotropic network with a width of 128, the implicit feature dimension number achieved through the star operation approximates $90^{1024}$, which can be reasonably approximated as infinite dimensions. 
Therefore, by stacking multiple layers, even just a few, star operations can substantially amplify the implicit dimensions in an exponential manner.

\subsection{Special Cases}
\label{sec:special_cases}
Not all star operations adhere to the formulation presented in Eq.~\ref{eq:star}, where each branch undergoes a transformation. For instance, VAN\cite{guo2023visual} and SENet~\cite{hu2018squeeze} incorporate one identity branch, while GENet-$\theta^-$~\cite{hu2018gather} operates without any learnable transformation. Subsequently, we will delve into the intricacies of these unique cases.

\noindent\textbf{Case I: Non-Linear Nature of $\mathbf{W_1}$ and/or $\mathbf{W_2}$}. In practical scenarios, a significant number of studies (e.g., Conv2Former, FocalNet, etc.) implement the transformation functions $\mathrm{W}_1$ and/or $\mathrm{W}_2$ as non-linear by incorporating activation functions. Nonetheless, a critical aspect is their maintenance of channel communications, as depicted in Eq.~\ref{eq:star_second}. Importantly, the number of implicit dimensions remains unchanged (approximately $\frac{d^2}{2}$), thereby not affecting our analysis in Sec.~\ref{sec:star_operation_in_one_layer}. Hence, we can simply use the linear transformation as a demonstration.  

\noindent\textbf{Case II: $\mathbf{W_1^TX *X}$}. When removing the transformation $\mathrm{W}_2$, the implicit dimension number decreases from approximately $\frac{d^2}{2}$ to $2d$.

\noindent\textbf{Case III: $\mathbf{X *X}$}.  In this case, star operation converts the feature from a feature space $\left\{ x^1, x^2, \cdots, x^d\right\}\in \mathbb{R}^d$ to a new space characterized by $\left\{ x^1x^1, x^2x^2, \cdots, x^dx^d\right\}\in \mathbb{R}^d$.

There are several notable aspects to consider. First, star operations and their special cases are commonly (though not necessarily) integrated with spatial interactions, typically implemented via pooling or convolution, as exemplified in VAN~\cite{guo2023visual}. Many of these approaches emphasize the benefits of an expanded receptive field, yet often overlook the advantages conferred by implicit high-dimensional spaces. Second, it is feasible to combine these special cases, as demonstrated in Conv2Former~\cite{hou2022conv2former}, which merges aspects of Case I and Case II, and in GENet-$\theta^-$~\cite{hu2018gather}, which blends elements of Case I and Case III. Lastly, although Cases II and III may not significantly increase the implicit dimensions in a single layer, the use of linear layers (primarily for channel communication) and skip connections can cumulatively achieve high implicit dimensions across multiple layers.

\begin{table}[]
\resizebox{\linewidth}{!}{
\begin{tabular}{lccccccc}
\toprule
Width & 96   & 128  & 160  & 192  & 224  & 256  & 288      \\
\midrule
sum acc.  & 51.1 & 58.5 & 62.7 & 66.2 & 69.6 & 71.4 & 72.5   \\
 star acc. & 57.6 & 64.0 & 68.2 & 71.7 & 73.9 & 75.3 & 76.1  \\
 \arrayrulecolor{gray}\hline
 acc. gap  & 6.5$\uparrow$  & 5.5$\uparrow$  & 5.5$\uparrow$  & 5.5$\uparrow$  & 4.3$\uparrow$  & 3.9$\uparrow$  & 3.6$\uparrow$   \\ 
 \arrayrulecolor{black}\bottomrule
\end{tabular}
}
\caption{\textbf{ImageNet-1k classification accuracy of DemoNet using sum operation or star operation with different widths.} We set the depth to 12. We gradually increase the width by a step of 32. }
\label{tab:demonet_width}
\end{table}

\begin{table}[]
\resizebox{\linewidth}{!}{
\begin{tabular}{lccccccc}
\toprule
Depth    & 10   & 12   & 14   & 16   & 18   & 20   & 22     \\
\midrule
sum acc.  & 63.8 & 66.3 & 68.2 & 68.8 & 69.9 & 69.6 & 70.6  \\
star acc. & 70.3 & 71.8 & 72.9 & 72.9 & 73.9 & 75.4 & 75.4  \\
\arrayrulecolor{gray}\hline
acc. gap  & 6.5$\uparrow$  & 5.5$\uparrow$  & 4.7$\uparrow$  & 4.1$\uparrow$  & 4.0$\uparrow$    & 5.8$\uparrow$  & 4.8$\uparrow$    \\
\arrayrulecolor{black}\bottomrule
\end{tabular}
}
\caption{\textbf{ImageNet-1k classification accuracy of DemoNet using sum operation or star operation with different depths.} We set the width to 192. We gradually increase the depth by a step of 2.}
\label{tab:demonet_depth}
\end{table}

\subsection{Empirical Study}
To substantiate and validate our analysis, we conduct extensive studies on the star operation from various perspectives. 
\subsubsection{Empirical superiority of star operation}
\label{sec:Empirical_superiority_of_star_operation}
Initially, we empirically validate the superiority of the star operation compared to simple summation. As illustrated in Fig.~\ref{fig:teaser}, we build an isotropic network, referred to as DemoNet, for this demonstration. DemoNet is designed to be straightforward, consisting of a convolutional layer that reduces the input resolution by a factor of 16, followed by a sequence of homogeneous demo blocks for feature extraction (refer to Fig.~\ref{fig:teaser}, left side). Within each demo block, we apply either the star operation or the summation operation to amalgamate features from two distinct branches. By varying the network's width and depth, we explore the distinctive attributes of each operation. The implementation details of DemoNet are provided in the supplementary Algorithm~\ref{alg: demonet_code}.

From Table~\ref{tab:demonet_width} and Table~\ref{tab:demonet_depth}, we can see that star operation consistently outperforms sum operation, regardless of the network depth and width. This phenomenon verifies the effectiveness and superiority of star operation. Moreover, we observed that with the increase in network width, the performance gains brought by the star operation gradually diminish. However, we did not observe a similar phenomenon in the case of varying depths. This disparity in behavior suggests two key insights: 1) The gradual decrease in the gains brought by the star operation as shown in Table~\ref{tab:demonet_width} is not a consequence of the model's enlarged size; 2) Based on this, it implies that the star operation does intrinsically expand the network’s dimensionality, which in turn lessens the incremental benefits of widening the network.

\begin{figure}
    \centering
    \includegraphics[width=1.0\linewidth]{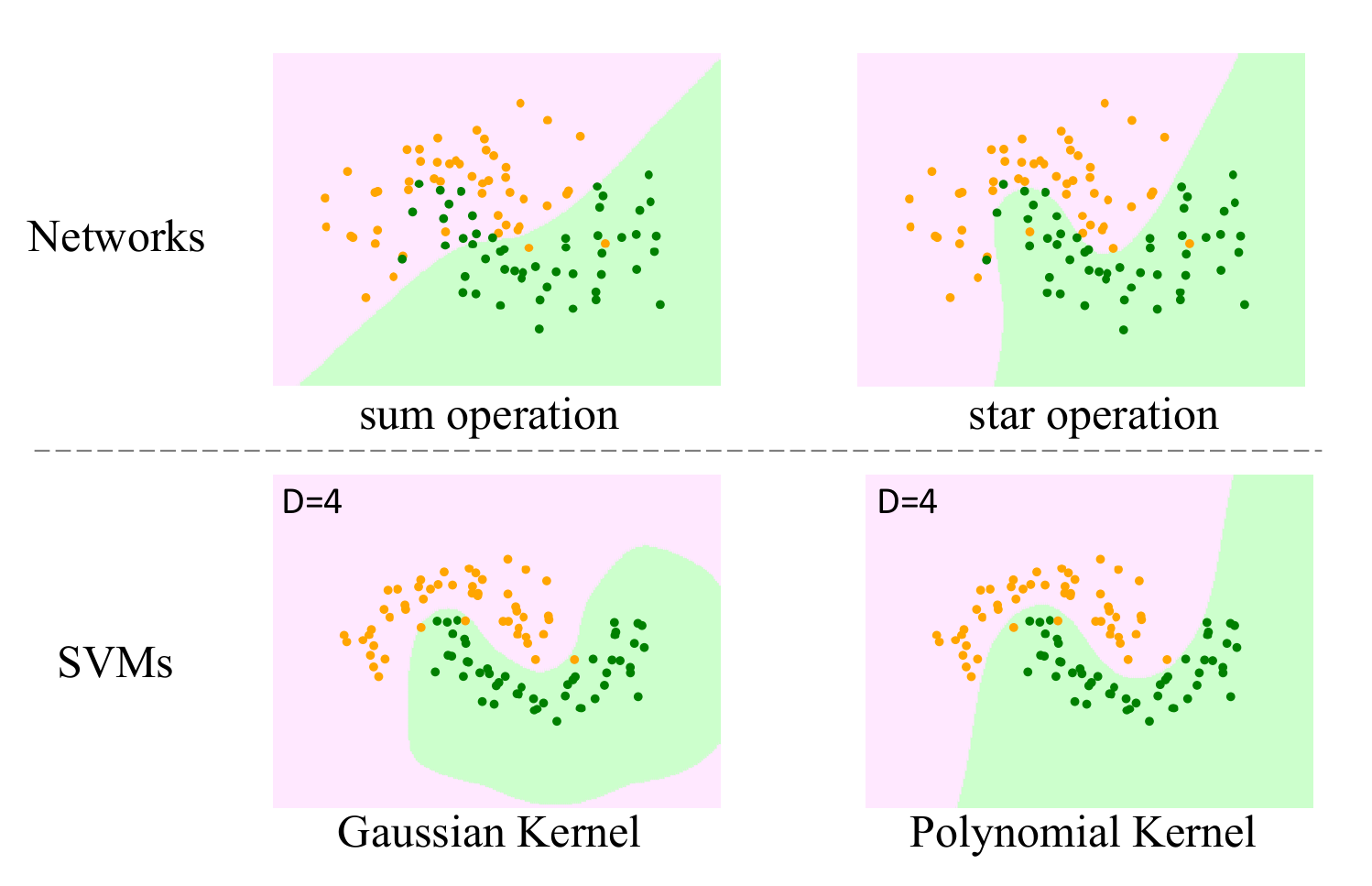}
    \caption{\textbf{Decision Boundary Comparison on 2D Noisy Moon Dataset}~\cite{scikit-learn}. The star-based network exhibits a more effective decision boundary than summation under identical configurations. Relative to SVMs, the star operation's boundary closely aligns with that of a polynomial kernel SVM, differing from the Gaussian kernel SVM. More details are available in the Supplementary.} 
    \label{fig:2d_points}
\end{figure}
\subsubsection{Decision Boundary comparison}
Subsequently, we visually analyze and discern the differences between the star and summation operations.  For this purpose, we visualize the decision boundaries of these two operations on the toy 2D moon dataset~\cite{scikit-learn}, which consists of two sets of moon-shaped 2D points.  In terms of the model configuration, we eliminate normalization and convolutional layers from the demo block. Given the relatively straightforward nature of this dataset, we configure the model with a width of 100 and a depth of 4.

Fig.~\ref{fig:2d_points} (top row) displays the decision boundaries delineated by the sum and star operations. It is evident that the star operation delineates a significantly more precise and effective decision boundary compared to the sum operation. Notably, the observed differences in decision boundaries do not stem from non-linearity, as both operations incorporate activation functions in their respective building blocks. The primary distinction arises from the star operation's capability to attain exceedingly high dimensionality, a characteristic we have previously analyzed in detail.

As aforementioned, the star operation functions analogously to kernel functions, particularly the polynomial kernel function. To corroborate this, we also illustrate the decision boundaries of SVM with both Gaussian and polynomial kernels (implemented using the scikit-learn package~\cite{scikit-learn}) in Fig.~\ref{fig:2d_points} (bottom row). In line with our expectations, the decision boundary produced by the star operation closely mirrors that of the polynomial kernel, while markedly diverging from the Gaussian kernel. This compelling evidence further substantiates the correctness of our analysis.

\begin{table}[]
\centering
\begin{tabular}{cccc}
\toprule
operation & w/ act. & w/o act. & Accuracy drop \\
\midrule
sum       & 66.2    & 32.4     & \textcolor{red}{33.8$\downarrow$}             \\
star      & 71.7    & 70.5     & \textcolor{purple}{1.2$\downarrow$}  \\
\bottomrule
\end{tabular}
\caption{
\textbf{DemoNet (width=192, depth=12) performance with and without activations}. Removing all activations leads to a significant performance drop in summation, whereas the star operation maintains its efficacy when removing ALL activations.}
\label{tab:without_activation}
\end{table}

\subsubsection{Extension to networks without activations}
Activation functions are fundamental and indispensable components in neural networks. Commonly employed activations like ReLU and GELU, however, are subject to certain drawbacks such as `mean shift'~\cite{brock2021high,hendrycks2016gaussian,he2015delving} and information loss~\cite{sandler2018mobilenetv2}, among others. The prospect of excluding activation functions from networks is an intriguing and potentially advantageous concept. Nevertheless, without activation functions, traditional neural networks would collapse into a single-layer network due to the lack of non-linearity.

In this study, while our primary focus is on the implicit high-dimensional feature achieved via star operations, the aspect of non-linearity also holds profound importance. To investigate this, we experiment by removing \textbf{all} activations from DemoNet, thus creating an activation-free network. 
The results in Table~\ref{tab:without_activation} are highly encouraging. As expected, the performance of the summation operation deteriorates markedly upon the removal of all activations, from 66.2\% to 32.4\%. In stark contrast, the star operation experiences only a minimal impact from the elimination of activations, evidenced by a mere 1.2\% decrease in accuracy. This experiment not only corroborates our theoretical analysis but also paves the way for expansive avenues in future research.

\begin{figure}[!t]
    \centering
    \includegraphics[width=1\linewidth]{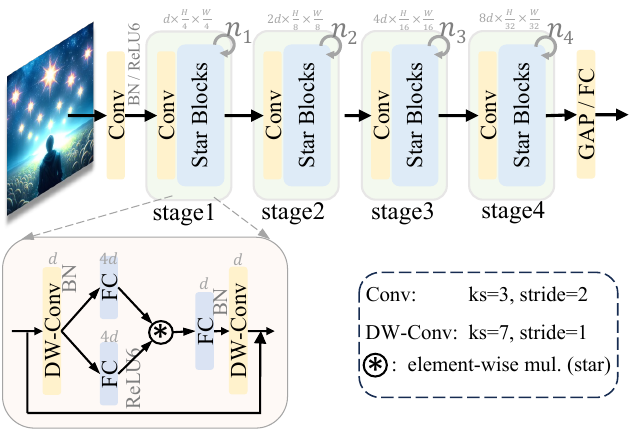}
    \caption{
    \textbf{StarNet architecture overview}. StarNet follows traditional hierarchical networks, and directly uses the convolutional layer to down-sample the resolution and double the channel number in each stage. We repeat multiple star blocks to extract features. Without any intricate structures and carefully chosen hyper-parameters, StarNet is able to deliver promising performance.}
    \label{fig:starnet_architecture}
\end{figure}

\subsection{Open Discussions \& Broader Impacts}
\label{sec:discussions}
Although based on the simple operation, our analysis lays the groundwork for exploring fundamental challenges in deep learning. Below, we outline several promising and intriguing research questions that merit further investigation, where the star operation could play a pivotal role.

\noindent\textcolor{black}{\textbf{I.} Are activation functions truly indispensable?} In our research, we have concentrated on the aspect of implicit high dimensions introduced by star operations. Notably, star operations also incorporate non-linearity, a characteristic that distinguishes kernel functions from other linear machine learning methods. A preliminary experiment in our study demonstrated the potential feasibility of eliminating activation layers in neural networks.

\noindent\textcolor{black}{\textbf{II.} How do star operations relate to self-attention and matrix multiplication?} Self-attention utilizes matrix multiplication to produce matrices in $\mathbb{R}^{n\times n}$. It can be demonstrated that matrix multiplication in self-attention shares similar attributes (non-linearity and high dimensionality) with element-wise multiplication. Notably, matrix multiplication facilitates global interactions, in contrast to the element-wise multiplication. However, matrix multiplication alters the input shape, necessitating additional operations (e.g., pooling, another round of matrix multiplication, etc.) to reconcile tensor shape, a complication avoided by element-wise multiplication. PolyNL~\cite{babiloni2021poly} provides preliminary efforts in this direction. Our analysis could offer new insights into the effectiveness of self-attention and contribute to the revisiting on `dynamic' features~\cite{chen2020dynamic,chen2020dynamicrelu,dai2021dynamic} in neural networks.

\noindent\textcolor{black}{\textbf{III.} How to optimize the coefficient distribution in implicit high-dimensional spaces?} Traditional neural networks can learn a distinct set of weight coefficients for each channel, but the coefficients for each implicit dimension in star operations, akin to kernel functions, are fixed. For instance, in the polynomial kernel function $k\left(x_1, x_2\right) = \left(\gamma x_1\cdot x_2 + c\right)^d$, the coefficient distribution can be adjusted via hyper-parameters. In star operations, while the weights $\mathrm{W}_1$ and $\mathrm{W}_2$ are learnable, they offer only limited scope for fine-tuning the distribution, as opposed to allowing for customized coefficients for each channel as in traditional neural networks. This constraint might explain why extremely high dimensions result in only moderate performance improvements. Notably, skip connections seem to aid in smoothing the coefficient distribution~\cite{veit2016residual}, and dense connections (as in DenseNet~\cite{huang2017densely}) may offer additional benefits. Furthermore, employing exponential functions could provide a direct mapping to implicit infinite dimensions, similar to Gaussian kernel functions.

\begin{table}[]
\centering
\begin{tabular}{l|cc|cc}
\toprule
Variant    & embed & depth             & Params & FLOPs \\
\midrule
StarNet-s1 & 24           & {[}2, 2, 8, 3{]}     & 2.9M    & 425M   \\
StarNet-s2 & 32              & {[}1, 2, 6, 2{]}  & 3.7M    & 547M   \\
StarNet-s3 & 32              & {[}2, 2, 8, 4{]}  & 5.8M    & 757M   \\
StarNet-s4 & 32              & {[}3, 3, 12, 5{]} & 7.5M    & 1075M \\
\bottomrule
\end{tabular}
\caption{
\textbf{Configurations of StarNets}.  We only vary the embed width and the depth to build different sizes of StarNet.}
\label{tab:starnet_configurations}
\end{table}

\section{Proof-of-Concept: StarNet}

Given the unique advantage of the star operation — its ability to compute in a low-dimensional space while yielding high-dimensional features — we identify its utility in the domain of efficient network architectures. Consequently, we introduce StarNet as a proof-of-concept model. StarNet is characterized by an exceedingly minimalist design and a significant reduction in human intervention. Despite its simplicity, StarNet showcases exceptional performance, underscoring the efficacy of the star operation.

\subsection{StarNet Architecture}
\label{sec:starnet_architecture}
StarNet is structured as a 4-stage hierarchical architecture, utilizing convolutional layer for down-sampling and a modified demo block for feature extraction. To meet the requirement of efficiency, we replace Layer Normalization with Batch Normalization and place it after the depth-wise convolution (can be fused during inference). Drawing inspiration from MobileNeXt\cite{zhou2020rethinking}, we incorporate a depth-wise convolution at the end of each block. The channel expansion factor is consistently set at 4, with network width doubling at each stage. The GELU activation in the demo block is substituted with ReLU6, following the MobileNetv2~\cite{sandler2018mobilenetv2} design. The StarNet framework is illustrated in Fig.~\ref{fig:starnet_architecture}. We only vary the block numbers and input embedding channel number to build different sizes of StarNet, as detailed in Table~\ref{tab:starnet_configurations}.

While many advanced design techniques (like re-parameterization, integrating with attention, SE-block, \textit{etc.}) can empirically enhance performance, but will also obscure our contributions. By deliberately eschewing these sophisticated design elements and minimizing human design intervention, we underscore the pivotal role of star operations in the conceptualization and functionality of StarNet.

\begin{table}[]
\centering
     \caption{\textbf{Comparison of Efficient Models on ImageNet-1k}. Models with a size under 1G FLOPs are compared, and sorted by parameter count. Latency is evaluated across various platforms, including Intel E5-2680 CPU, P100 GPU, and iPhone 13 mobile device. Latency benchmarking batch size is set to 1 as in real-world scenario.}
    \label{tab:compare_imagenet}
    \resizebox{\linewidth}{!}{
    \begin{tabular}{p{3cm}p{0.9cm}<{\centering}p{0.8cm}<{\centering}p{0.8cm}<{\centering}p{0.7cm}<{\centering}p{0.6cm}<{\centering}p{0.6cm}<{\centering}}
        \toprule
    \multirow{2}{*}{\textbf{Model}} &\textbf{Top-1}& {\textbf{Params}} & {\textbf{FLOPs}} & \multicolumn{3}{c}{\bf Latency (ms)} \\
    \cmidrule{5-7}
        &\textbf{(\%)} & \textbf{(M)} & \textbf{(M)} & \textbf{Mobile} & \textbf{GPU} &\textbf{CPU} \\
        \toprule
        MobileOne-S0~\cite{vasu2023mobileone}     & 71.4  & 2.1    & 275   &\textbf{0.7} 	&1.1	&2.2     \\
        ShuffleV2-1.0~\cite{ma2018shufflenet} & 69.4  & 2.3    & 146  &4.1	&2.2	  & 3.8     \\
        MobileV3-S0.75~\cite{howard2019searching}& 65.4  & 2.4    & 44   &5.5 	&1.8	 &2.5       \\
        GhostNet0.5~\cite{han2020ghostnet}      & 66.2  & 2.6    & 42    &10.0 	&2.9	 &4.8   \\
        MobileV3-S~\cite{howard2019searching}    & 67.4  & 2.9    & 66   &6.5 	&1.8	 &2.6     \\
        StarNet-S1	&\textbf{73.5}	&2.9	&425	&\textbf{0.7}	&2.3 &4.3	 \\
        \midrule  %
        MobileV2-1.0~\cite{sandler2018mobilenetv2} & 72.0  & 3.4    & 300   &0.9	&2.2	 & 3.2       \\
        ShuffleV2 1.5~\cite{ma2018shufflenet} & 72.6  & 3.5    & 299   &5.9	&2.2	& 4.9     \\
        Mobileformer-52~\cite{chen2022mobile}  & 68.7  & 3.6    & 52   &6.6  	&8.3 & 26.0	     \\
        FasterNet-T0~\cite{chen2023run}     & 71.9  & 3.9    & 338  &\textbf{0.7} 	&2.5	   & 5.7    \\
        
        StarNet-S2	&\textbf{74.8}	&3.7	&547	&\textbf{0.7}	&2.0&4.5	 \\
        \midrule  %
        MobileV3-L0.75~\cite{howard2019searching}& 73.3  & 4.0    & 155  &10.9  	&2.2	  &4.4  \\
        EdgeViT-XXS~\cite{pan2022edgevits}      & 74.4  & 4.1    & 559   &1.8	&8.9	 &12.6         \\
        MobileOne-S1~\cite{vasu2023mobileone}     & 75.9  & 4.8    & 825  &\textbf{0.9}	&1.5	  &6.0 \\
        GhostNet1.0~\cite{han2020ghostnet}     & 73.9  & 5.2    & 141   &7.9 	&3.6	   &7.0  \\
        EfficientNet-B0~\cite{tan2019efficientnet}  & 77.1  & 5.3    & 390  &1.6    	&3.4	&8.8   \\
        MobileV3-L~\cite{howard2019searching}    & 75.2  & 5.4    & 219 &11.4  	&2.5 &5.2	\\
        StarNet-S3	&\textbf{77.3}	&5.8	&757	&\textbf{0.9}	&2.7	&6.7\\
        \midrule  %
        EdgeViT-XS~\cite{pan2022edgevits}       & 77.5  & 6.8    & 1166 &3.5 &12.1	 &18.3	         \\
        MobileV2-1.4~\cite{sandler2018mobilenetv2} & 74.7  & 6.9    & 585  &1.1   	&2.8	 &5.4      \\
        GhostNet1.3~\cite{han2020ghostnet}     & 75.7  & 7.3    & 226  &9.7  	&3.9	     &11.0    \\
        ShuffleV2-2.0~\cite{ma2018shufflenet} & 74.9  & 7.4    & 591  &19.9	&2.6	 & 9.7          \\
        Fasternet-T1~\cite{chen2023run}     & 76.2  & 7.6    & 855   &\textbf{0.9} 	&3.3	   & 9.7     \\
        MobileOne-S2~\cite{vasu2023mobileone}     & 77.4  & 7.8    & 1299 &\textcolor{darkgray}{\textbf{1.0}}  	&2.0	 & 8.9     \\
        StarNet-S4	&\textbf{78.4}	&7.5	&1075 &\textcolor{darkgray}{\textbf{1.0}} 		&3.7	&9.4\\
\bottomrule
    \end{tabular}
    }
\end{table}

\subsection{Experimental Results}
We adhere to a standard training recipe from DeiT~\cite{touvron2021training} to ensure a fair comparison when training our StarNet models. All models are trained from scratch over 300 epochs, utilizing the AdamW optimizer~\cite{loshchilov2017decoupled} with an initial learning rate of 3e-3 and a batch size of 2048. Comprehensive training details are provided in the supplementary materials. For benchmark purposes, our PyTorch models are converted to the ONNX format~\cite{onnxruntime} to facilitate latency evaluations on both CPU (Intel Xeon CPU E5-2680 v4 @ 2.40GHz) and GPU (P100). Additionally, we deploy the models on iPhone13 using CoreML-Tools~\cite{coreml} to assess latency on mobile devices. Detailed settings for these benchmarks are also available in the supplementary materials.

\begin{figure}
    \centering
    \includegraphics[width=1\linewidth]{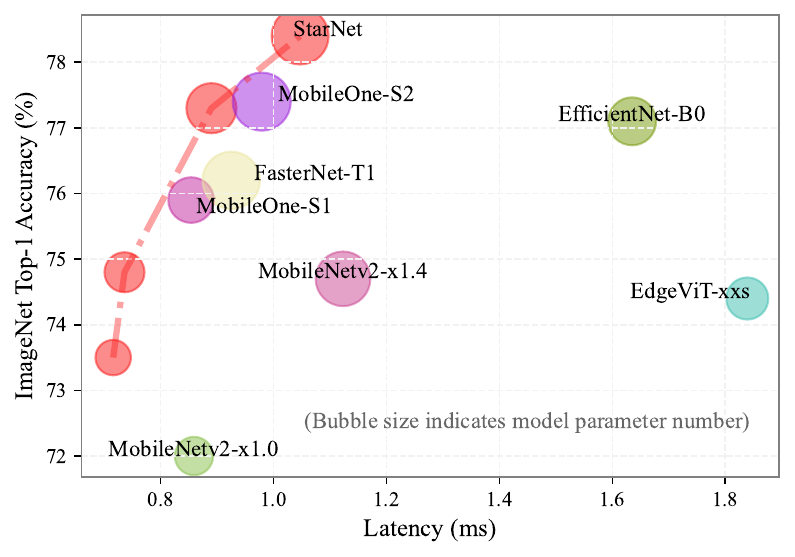}
    \caption{\textbf{Mobile Device (iPhone13) Latency vs. ImageNet Accuracy}. Models with excessively high latency are excluded from this figure. More results on different mobile devices can be found in supplementary Table~\ref{tab:more_iphone_latency}.}
    \label{fig:latency_accuracy}
\end{figure}

The experimental results are presented in Table~\ref{tab:compare_imagenet}. With minimal handcrafted design, our StarNet is able to deliver promising performance in comparison with many other state-of-the-art efficient models. Notably, StarNet achieves a top-1 accuracy of 73.5\% in just 0.7 seconds on iPhone 13 device, surpassing MobileOne-S0 by 2.1\% (73.5\% \textit{vs.} 71.4\%) at the same latency. When scaling the model to a 1G FLOPs budget, StarNet continues to exhibit remarkable performance, outperforming MobileOne-S2 by 1.0\%, and surpassing EdgeViT-XS by 0.9\% while being three times faster (1.0 ms vs. 3.5 ms).
This impressive efficiency, given the model's straightforward design, can be mainly attributed to the fundamental role of the star operation. Fig.~\ref{fig:latency_accuracy} further illustrates the latency-accuracy trade-off among various models.
If we can further push the performance of StarNet to a higher level? We believe that through careful hyper-parameter optimization, leveraging insights from Table~\ref{tab:summary_efficient_networks}, and applying training enhancements such as more epochs or distillation, substantial improvements could be made to StarNet's performance. However, achieving a high-performance model is not our primary goal, as such enhancements could potentially obscure the core contributions of the star operation.  We left the engineering work behind.

\subsection{More Ablation studies}
\paragraph{Substituting the star operation.} The star operation is identified as the sole contributor to the high performance of our model. To empirically validate this assertion, we systematically replaced the star operation with summation in our implementation. Specifically, this entailed substituting the `*' operator with a `+' in the model's architecture.

The results are delineated in Table~\ref{tab:replace_star}. Removing all star operations resulted in a notable performance decline, with a 3.1\% accuracy drop observed. Interestingly, the impact of the star operation on performance appears minimal in the first and second stages of the model. This observation is logical. With a very narrow width, the ReLU6 activation results in some features turning to zero. In the context of the star operation, this leads to numerous dimensions in its implicit high-dimensional space also becoming zero, thereby restraining its full potential. However, its contribution becomes more pronounced in the last two stages (more channels), leading to 1.6\% and 1.6\% improvements, respectively. Last three rows in Table~\ref{tab:replace_star} also validate our analysis.

\paragraph{Latency impact of the star operation.} Theoretically, multiplication operations (such as the star operation in our study) are understood to have a higher computational complexity compared to simpler summation operations, as indicated in several related works~\cite{chen2020addernet,elhoushi2021deepshift}. However, practical latency outcomes may not always align with theoretical predictions. We conducted benchmarks to compare the latency of replacing all star operations with summation, with the results detailed in Table~\ref{tab:latency_star_vs_sum}. 
From the table, we observed that the latency impact is contingent on the hardware. In practice, the star operation did not result in any additional latency on GPU and iPhone devices relative to the summation operation. However, the summation operation was slightly more efficient than the star operation on CPU (\textit{e.g.}, 8.4ms vs. 9.4 ms for StarNet-S4). Given the considerable performance gap, this minor latency overhead on CPU can be deemed negligible.

\begin{table}[]
\centering

\begin{tabular}{ccccc}
\toprule
Stage 1 & Stage 2 & Stage 3 & Stage 4 & Top-1 \\
\midrule
\cellcolor{LightGreen}sum       & \cellcolor{LightGreen}sum      & \cellcolor{LightGreen}sum      & \cellcolor{LightGreen}sum       &    75.3  \\
\cellcolor{LightBlue}star       & \cellcolor{LightGreen}sum       & \cellcolor{LightGreen}sum     &  \cellcolor{LightGreen}sum       &   75.1   \\
\cellcolor{LightBlue}star      & \cellcolor{LightBlue}star      & \cellcolor{LightGreen}sum       & \cellcolor{LightGreen}sum       &     75.2 \\
\cellcolor{LightBlue}star      & \cellcolor{LightBlue}star       & \cellcolor{LightBlue}star       & \cellcolor{LightGreen}sum       &   76.8   \\
\cellcolor{LightBlue}star       & \cellcolor{LightBlue}star       & \cellcolor{LightBlue}star       & \cellcolor{LightBlue}star       &    \textbf{78.4}  \\
\midrule
\cellcolor{LightGreen}sum       & \cellcolor{LightGreen}sum        & \cellcolor{LightBlue}star       & \cellcolor{LightGreen}sum        &    76.4  \\
\cellcolor{LightGreen}sum        & \cellcolor{LightGreen}sum       & \cellcolor{LightGreen}sum       & \cellcolor{LightBlue}star       &    76.9  \\
\cellcolor{LightGreen}sum        & \cellcolor{LightGreen}sum        & \cellcolor{LightBlue}star       & \cellcolor{LightBlue}star       &    \textbf{78.4}  \\
\bottomrule
\end{tabular}
\caption{\textbf{Gradually replacing star operation `*' with summation `+' in StarNet-S4} (considering its sufficient depth and model size).}
\label{tab:replace_star}
\end{table}

\begin{table}[]\centering

\begin{tabular}{cccl}
\toprule
StarNet & \makecell{Mobile (ms) \\ {\footnotesize sum / star}} & \makecell{GPU (ms) \\ {\footnotesize sum / star}} & \makecell{CPU (ms) \\ {\footnotesize sum / star}}\\
\midrule
S1  & 0.7 /  0.7 {\tiny \textcolor{gray}{(------)}} & 2.3 / 2.3 {\tiny \textcolor{gray}{(------)}} & 3.8 / 4.3 {\tiny \textcolor{magenta}{($+$0.5)}}\\

S2   & 0.8 / 0.7 {\tiny \textcolor{cyan}{($-$ 0.1)}} & 2.0 / 2.0 {\tiny \textcolor{gray}{(------)}}& 4.2 / 4.5 {\tiny \textcolor{magenta}{($+$0.3)}} \\
 
S3  & 0.9 / 0.9 {\tiny \textcolor{gray}{(------)}}  & 2.7 / 2.7 {\tiny \textcolor{gray}{(------)}} & 5.9 / 6.7 {\tiny \textcolor{magenta}{($+$0.8)}}\\
S4   & 1.1 / 1.0 {\tiny \textcolor{cyan}{($-$ 0.1)}} & 3.7 / 3.7 {\tiny \textcolor{gray}{(------)}}  & 8.4 / 9.4 {\tiny \textcolor{magenta}{($+$1.0)}}\\
 \bottomrule
\end{tabular}

\caption{\textbf{Latency comparison of different operations in StarNet.}} 
\label{tab:latency_star_vs_sum}
\end{table}

\paragraph{Study on the activation placement.}  We present a comprehensive analysis regarding the placement of activation functions (ReLU6) within our network block. For clarity, $x_1$ and $x_2$ are used to denote the outputs of the two branches, with StarNet-S4 serving as the demonstrative model.

Here, we investigated four approaches to implementing activation functions within StarNet: 1) employing no activations, 2) activating both branches, 3) activating post star operation, and 4) activating a single branch, which is our default practice. The results, as depicted in Table~\ref{tab:activation_placement}, demonstrate that activating only one branch yields the highest accuracy, reaching 78.4\%. \textit{Inspiringly, the complete removal of activations from StarNet (\textcolor{darkgray}{except one in the stem layer}) leads to a mere 2.8\% reduction in accuracy, bringing it down to 75.6\%, a performance that is still competitive with some strong baselines in Table~\ref{tab:compare_imagenet}}. These findings, consistent with Table~\ref{tab:without_activation}, underscore the potential of activation-free networks.

\begin{table}[]
    \centering

    \begin{tabular}{cccc}
    \toprule
     $x_1$*$x_2$  & \textcolor{magenta}{act(}$x_1$\textcolor{magenta}{)}*\textcolor{magenta}{act(}$x_2$\textcolor{magenta}{)} & \textcolor{magenta}{act(}$x_1$*$x_2$\textcolor{magenta}{)} & \textcolor{magenta}{act(}$x_1$\textcolor{magenta}{)}*$x_2$\\
     {\footnotesize\textcolor{gray}{(no act.)}} & {\footnotesize\textcolor{gray}{(act. both)}} & {\footnotesize\textcolor{gray}{(post act.)}} & {\footnotesize\textcolor{gray}{(act. one)}} \\
    \midrule
     75.6& 78.0 & 77.0&\textbf{78.4} \\
    \bottomrule
    \end{tabular}
    
    \caption{\textbf{Results of diverse activation placements in StarNet-S4.}}
    \label{tab:activation_placement}
\end{table}

\paragraph{Study on the design of block with star operation.} In StarNet, the star operation is typically implemented as $\mathrm{act}\left(\mathrm{W}_1^\mathrm{T}\mathrm{X}\right) * \left(\mathrm{W}_2^\mathrm{T}\mathrm{X}\right)$, as detailed in Sec.~\ref{sec:star_operation_in_one_layer}. This standard approach allows StarNet-S4 to reach an accuracy of 84.4\%. However, alternative implementations are possible. We experimented with a variation: $\left(\mathrm{W}_2^\mathrm{T}\mathrm{act}\left(\mathrm{W}_1^\mathrm{T}\mathrm{X}\right)\right)*\mathrm{X}$, where $\mathrm{W}_1\in\mathbb{R}^{d\times d'}$ is designed to expand the width, and $\mathrm{W}_2\in\mathbb{R}^{d'\times d}$ restores it back to $d$. This adjustment results in the transformation of only one branch, while the other remains unaltered. We vary $d'$ to ensure the  same computational complexity as StarNet-S4.
By doing so, the performance dropped from 78.4\% to 74.4\%. While a better and careful design might mitigate this performance gap (see supplementary), the marked difference in accuracy emphasizes the efficacy of our initial implementation in harnessing the star operation's capabilities,  and underscores the critical importance of transforming both branches in star operation.

\section{Conclusion}
In this study, we have delved into the intricate details of the star operation, going beyond the intuitive and plausible explanations as in previous research. We recontextualized the star operations, uncovering that their strong representational capacity is derived from implicitly high-dimensional spaces. In many ways, the star operation mirrors the behavior of polynomial kernel functions. Our analysis was rigorously validated through empirical, theoretical, and visual methods.  
Our results were mathematically and theoretically solid, aligning coherently with the analysis we have presented.
Building on this foundation, we positioned the star operation within the realm of efficient network designs and introduced StarNet, a simple prototype network. StarNet’s impressive performance, achieved without the reliance on sophisticated designs or meticulously chosen hyper-parameters, stands as a testament to the efficacy of star operations. Furthermore, our exploration of the star operation opens up numerous potential research avenues, as we discussed above.

{
    \small
    \bibliographystyle{ieeenat_fullname}
    \bibliography{ref}
}

\clearpage
\setcounter{page}{1}
\maketitlesupplementary
\setcounter{section}{0}
\renewcommand{\thesection}{\Alph{section}}

This supplementary document elaborates on the implementation details of DemoNet, as showcased in Figure~\ref{fig:teaser}, Table~\ref{tab:demonet_depth}, and Table~\ref{tab:demonet_width}. It also covers the simple network used in visualizing the decision boundary, illustrated in Figure~\ref{fig:2d_points}, and the implementation of our StarNet presented in Table~\ref{tab:compare_imagenet} (details can be found in Section~\ref{sec:supp_implementation_details}). Moreover, we provide a more granular view of decision boundary visualization in Section~\ref{sec:supp_decision_boundary_visualization}. Section~\ref{sec:supp_exploring_extremely_small_models} delves into our exploratory studies on ultra-compact models.The analysis of activations is presented in Sec.~\ref{sec:supp_activation_analysis}. Additionally, a detailed examination of block design is discussed in Section~\ref{sec:supp_block_analysis}.

\section{Implementation Details}
\label{sec:supp_implementation_details}

\subsection{Model Architecture}
\label{sec:supp_model_architecture}
\paragraph{DemoNet} In our isotropic DemoNet, featured in Fig.~\ref{fig:teaser}, Table~\ref{tab:demonet_depth}, and Table~\ref{tab:demonet_width}, detailed implementation is provided in Algorithm~\ref{alg: demonet_code}. We adjust the depth or width values to facilitate the experiments showcased in the aforementioned figures and tables.

\begin{algorithm}[H]\small
    \caption{PyTorch codes of network illustrated in Fig.~\ref{fig:teaser}.}
    \label{alg: demonet_code}
    \definecolor{codeblue}{rgb}{0.25,0.5,0.5}
    \definecolor{codepink}{rgb}{1,0.5,0.5}
    \definecolor{codedark}{rgb}{1,0.7,0.8}
    \lstset{
        backgroundcolor=\color{white},
        basicstyle=\fontsize{7.2pt}{7.2pt}\ttfamily\selectfont,
        columns=fullflexible,
        breaklines=true,
        captionpos=b,
        commentstyle=\fontsize{8pt}{8pt}\color{codeblue},
        keywordstyle=\fontsize{8.0pt}{8.0pt}\color{codepink},
        emph={DemoNet, Blk}, %
        emphstyle=\color{purple}, %
    }
    {\small
    \begin{lstlisting}[language=python]
# demo-code for our DemoNet in Fig. 1.
# We build an isotropic network with depth 12. 
# params: dim: width of network; mod: "star" or "sum".

import torch.nn as nn
from torch.nn import Sequential as Seq
import torch.nn.functional as F  

class Blk(nn.Module):
  def __init__(self, dim, mod="sum"):
    super().__init__()
    self.mod=mod
    self.norm = nn.LayerNorm(dim)
    self.dwconv = nn.Conv2d(dim,dim,7,1,3,groups=dim)
    self.f = nn.Linear(dim, 6 * dim)
    self.g = nn.Linear(3*dim, dim)
    
    def forward(self, x):
      input = x
      x = self.dwconv(self.norm(x).permute(0,3,1,2))
      x = self.f(x.permute(0,2,3,1))
      x1, x2 = x.split(x.size(-1)//2, dim=-1)
      x = F.gelu(x1)+x2 if self.mod == "sum" \\
          else F.gelu(x1)*x2
      x = self.g(x)
      x = input + x
      return x

class DemoNet(nn.Module):
  def __init__(self, dim=128, mod="star", depth=12):
    super().__init__()
    self.num_classes = 1000
    self.stem = nn.Conv2d(3, dim, 16, 16)
    self.net=Seq(*[Blk(dim,mod) for _ in range(depth)])
    self.norm = nn.LayerNorm(dim) 
    self.head = nn.Linear(dim, self.num_classes)

  def forward(self, x):
    x = self.net(self.stem(x).permute(0,2,3,1))
    return self.head(self.norm(x.mean([1, 2])))
    \end{lstlisting}
    }
\end{algorithm}

\paragraph{DemoNet for 2D Points} For the illustration of 2D points, as shown in Fig.~\ref{fig:2d_points}, we have further simplified the DemoNet structure. In this adaptation, all convolutional layers within the original DemoNet have been removed. Additionally, we replaced the GELU activation function with ReLU to streamline the architecture. The details of this simplified DemoNet are outlined in Algorithm~\ref{alg: demonet2d_code}.

\begin{algorithm}[H]\small
    \caption{PyTorch codes of network illustrated in Fig.~\ref{fig:2d_points}.}
    \label{alg: demonet2d_code}
    \definecolor{codeblue}{rgb}{0.25,0.5,0.5}
    \definecolor{codepink}{rgb}{1,0.5,0.5}
    \definecolor{codedark}{rgb}{1,0.7,0.8}
    \lstset{
        backgroundcolor=\color{white},
        basicstyle=\fontsize{7.2pt}{7.2pt}\ttfamily\selectfont,
        columns=fullflexible,
        breaklines=true,
        captionpos=b,
        commentstyle=\fontsize{8pt}{8pt}\color{codeblue},
        keywordstyle=\fontsize{8.0pt}{8.0pt}\color{codepink},
        emph={DemoNet2D, Blk}, %
        emphstyle=\color{purple}, %
    }
    {\small
    \begin{lstlisting}[language=python]
# demo-code for Network in Fig. 2.
# We build an isotropic network with depth 4, width 100 
# params: mod: "star" or "sum".

import torch.nn as nn
from torch.nn import Sequential as Seq
import torch.nn.functional as F  

class Blk(nn.Module):
  def __init__(self, dim, mod="sum"):
    super().__init__()
    self.mod=mod
    self.norm = nn.LayerNorm(dim)
    self.f = nn.Linear(dim, 6 * dim)
    self.g = nn.Linear(3*dim, dim)
    
    def forward(self, x):
      input = x
      x = self.f(self.norm(x))
      x1, x2 = x.split(x.size(-1)//2, dim=-1)
      x = F.relu(x1)+x2 if self.mod == "sum" \\
          else F.relu(x1)*x2
      x = self.g(x)
      x = input + x
      return x

class DemoNet2D(nn.Module):
  def __init__(self, dim=100, mod="star", depth=4):
    super().__init__()
    self.num_classes = 2
    self.stem = nn.Linear(2, dim)
    self.net=Seq(*[Blk(dim,mod) for _ in range(depth)])
    self.norm = nn.LayerNorm(dim) 
    self.head = nn.Linear(dim, self.num_classes)

  def forward(self, x):
    x = self.net(self.stem(x))
    return self.head(self.norm(x))
    \end{lstlisting}
    }
\end{algorithm}

\paragraph{StarNet} For ease of reproduction, we include a separate file in the supplementary materials dedicated to our StarNet. Detailed information regarding its architecture is also available in Section~\ref{sec:starnet_architecture}.

\subsection{Training Recipes}
\label{sec:supp_training_recipes}
We next provide detailed training recipe for each experiment.

\paragraph{DemoNet} For all DemoNet variants, we utilize a consistent and standard training recipe. While it's acknowledged that specialized and finely-tuned training recipes could better suit different model sizes and potentially yield enhanced performance, as in the cases of \textcolor{codedark}{DemoNet(width=96, depth=12)} (approximately 1.26M parameters) and \textcolor{codedark}{DemoNet(width=288, depth=12)} (approximately 9.68M parameters), achieving superior performance with DemoNet is not the primary goal of this work. Our objective is to provide a fair comparison among the various DemoNet variants; hence, the same training recipe is applied across all. Details of this training recipe for DemoNet are presented in Table~\ref{tab:supp_recipe_demonet}.

\begin{table}[!h]
\centering
\begin{tabular}{l@{\extracolsep{4pt}}!{\vrule width 1.0pt}c}
config & value \\
\Xhline{4\arrayrulewidth}
image size & 224 \\
optimizer & AdamW~\cite{loshchilov2017decoupled} \\
base learning rate & 4e-3 \\
weight decay & 0.05 \\
optimizer momentum & $\beta_1, \beta_2{=}0.9, 0.999$ \\
batch size & 2048 \\
learning rate schedule & cosine decay~\cite{loshchilov2016sgdr} \\
warmup epochs & 5 \\
training epochs & 300 \\
AutoAugment & rand-m9-mstd0.5-inc1 \cite{cubuk2018autoaugment} \\
label smoothing \cite{muller2019does} & 0.1 \\
mixup \cite{zhang2017mixup} & 0.8 \\
cutmix \cite{yun2019cutmix} & 1.0 \\
color jitter & 0.4 \\
drop path \cite{huang2016deep} & 0. \\
AMP & False \\
ema  & 0.9998 \\
\end{tabular}
\caption{\textbf{DemoNet training setting.
}}
\label{tab:supp_recipe_demonet}
\end{table}

\paragraph{DemoNet for 2D Points} Given the inherent simplicity of 2D points, we have eliminated all data augmentation processes and reduced the number of training epochs. The specific training recipe employed for this streamlined process is detailed in Table~\ref{tab:supp_recipe_demonet2d}.

\begin{table}[!h]
\centering
\begin{tabular}{l@{\extracolsep{4pt}}!{\vrule width 1.0pt}c}
config & value \\
\Xhline{4\arrayrulewidth}
optimizer & SGD\\
base learning rate & 0.1 \\
minimal learning rate & 0.005 \\
learning rate schedule & cosine decay~\cite{loshchilov2016sgdr} \\
weight decay & 2e-4 \\
optimizer momentum & 0.9 \\
batch size & 32 \\
training epochs & 30 \\
\end{tabular}
\caption{\textbf{Simplified DemoNet for 2D points training setting.
}}
\label{tab:supp_recipe_demonet2d}
\end{table}

\paragraph{StarNet} Owing to its small model size and straightforward architectural design, StarNet necessitates fewer augmentations for training regularization. \textbf{Importantly, we opted not to use the commonly employed Exponential Moving Average (EMA) and learnable layer-scale technique}~\cite{touvron2021going}. While these methods could potentially enhance performance, they may obscure the unique contributions of our work. The detailed training recipe for various StarNet models is provided in Table~\ref{tab:supp_recipe_starnet}.

\begin{table}
\centering
\begin{tabular}{l@{\extracolsep{4pt}}!{\vrule width 1.0pt}cccc}
config & value \\
\Xhline{4\arrayrulewidth}
image size & 224 \\
optimizer & AdamW~\cite{loshchilov2017decoupled} \\
base learning rate & 3e-3 \\
weight decay & 0.025 \\
optimizer momentum & $\beta_1, \beta_2{=}0.9, 0.999$ \\
batch size & 2048 \\
learning rate schedule & cosine decay~\cite{loshchilov2016sgdr} \\
warmup epochs & 5 \\
training epochs & 300 \\
AutoAugment & rand-m1-mstd0.5-inc1 \cite{cubuk2018autoaugment} \\
label smoothing \cite{muller2019does} & 0.1 \\
mixup \cite{zhang2017mixup} & 0.8 \\
cutmix \cite{yun2019cutmix} & 0.2 \\
color jitter & 0. \\
drop path \cite{huang2016deep} & 0.(S1/S2/S3), 0.1(S4) \\
\Xhline{4\arrayrulewidth}
AMP & False \\
EMA & None \\
layer-scale & None \\
\end{tabular}
\caption{\textbf{StarNet variants training setting.
}}
\label{tab:supp_recipe_starnet}
\end{table}

\subsection{Latency Benchmark Settings}

All models listed in Table~\ref{tab:compare_imagenet} have been converted from Pytorch code to the ONNX format~\cite{onnxruntime}, to enable latency evaluations on different hardware: CPU (Intel Xeon CPU E5-2680 v4 @ 2.40GHz) and GPU (P100). We have conducted benchmarks with a batch size of one, mirroring real-world application scenarios. The benchmark involves a warm-up of 50 iterations, followed by the computation of the average latency over 500 iterations. It's important to note that all models were benchmarked on the same device to ensure fairness in comparisons.
For CPU benchmarks, we utilize 4 threads to optimize performance. In the case of GPU evaluations, we've adapted the pointwise convolutional layer in StarNet to a linear layer, incorporating a permutation operation. This modification was prompted by observations of marginally faster inference speeds. It's important to note that, mathematically, a pointwise convolutional layer is equivalent to a linear layer. Therefore, this change does not lead to any variances in performance or alterations in the architecture of our StarNet.

Thanks to MobileOne~\cite{vasu2023mobileone}, we have utilized their open-sourced iOS benchmark application for all CoreML models. Our latency benchmark settings follow those used in MobileOne, with the only difference being the inclusion of additional models in our tests. We have observed that the initial run of the iOS benchmark consistently yields slightly faster results. To account for this, each model is run three times, and we report the latency of the final run. Although there are minor latency variations when testing on iPhone, these discrepancies (typically less than 0.05 ms) do not affect our analysis.

\label{sec:supp_latency_benchmark_settings}
\section{Decision Boundary Visualization}
\label{sec:supp_decision_boundary_visualization}
We provide more analyses for the decision boundary visualization, as shown in Fig.~\ref{fig:2d_points}.

\begin{figure}
    \centering
    \includegraphics[width=1.0\linewidth]{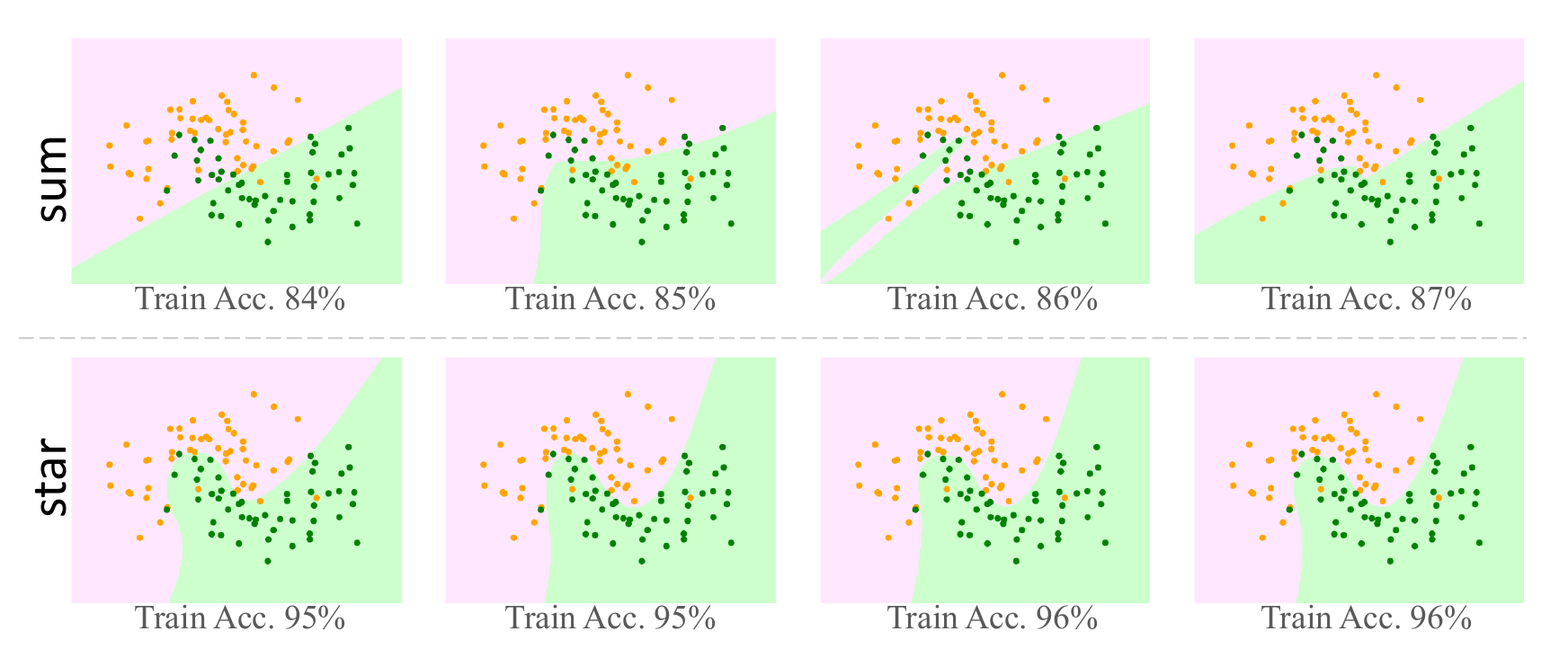}
    \caption{4-run results of sum and star operations.}
    \label{fig:2dpoints_4runs}
\end{figure}

Firstly, as a supplement to Fig.~\ref{fig:2d_points}, we present more comprehensive results. The decision boundary can exhibit significant variation due to randomness in different tests. To illustrate this, we display the decision boundaries from an additional four runs (with no fixed seed) in Fig.~\ref{fig:2dpoints_4runs}. As evidenced, the star operation not only surpasses the summation operation in representational capability, but it also demonstrates greater robustness, exhibiting minimal variance.

\begin{figure}[]
    \centering
    \begin{subfigure}[b]{1\linewidth}
        \includegraphics[width=\linewidth]{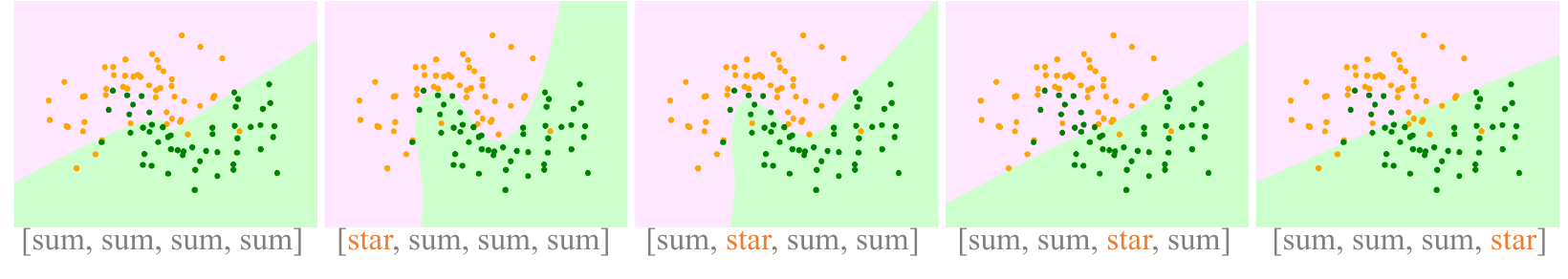}
        \caption{Replace sum with star for single block, from first to last.}
        \label{fig:supp_2dpoints_row1}
    \end{subfigure}
    \quad %
    \begin{subfigure}[b]{1\linewidth}
        \includegraphics[width=\linewidth]{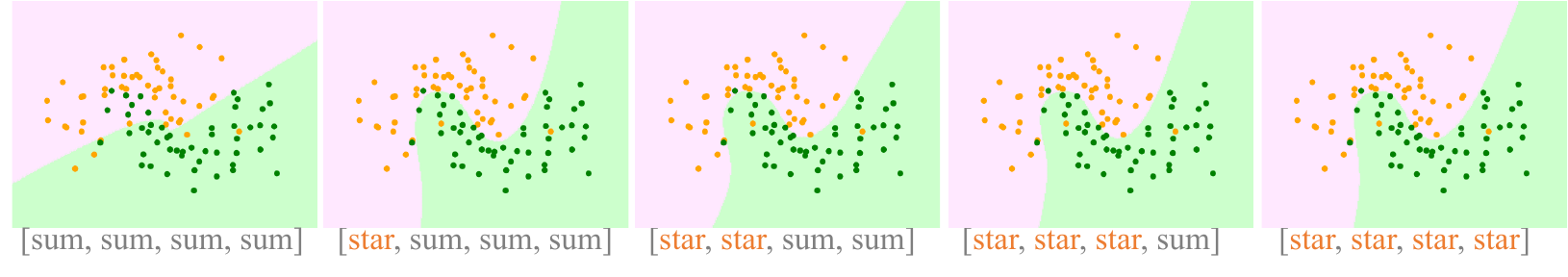}
        \caption{Progressively replace sum with star, from first to last.}
        \label{fig:supp_2dpoints_row2}
    \end{subfigure}
    
    \begin{subfigure}[b]{1\linewidth}
        \includegraphics[width=\linewidth]{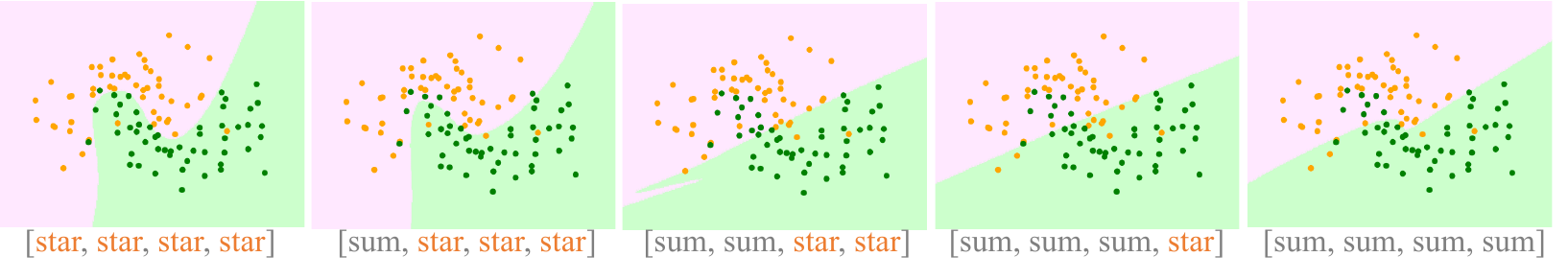}
        \caption{Progressively replace star with sum, from first to last.}
        \label{fig:supp_2dpoints_row3}
    \end{subfigure}
    
    \caption{More comprehensive analyses on decision boundary.}
    \label{fig:supp_2dpoints_comprehensive}
\end{figure}

\begin{figure}[!h]
    \centering
    \vspace{-4mm}
    \includegraphics[width=1\linewidth]{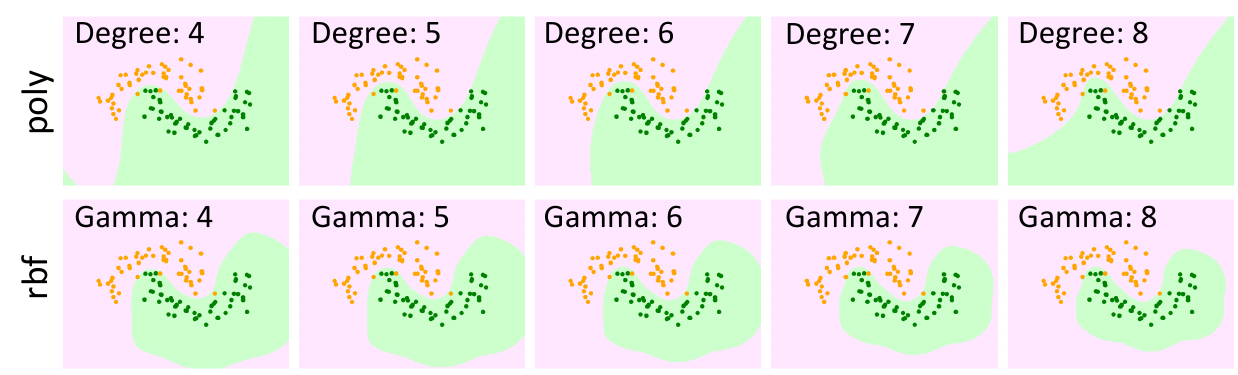}
    \vspace{-6mm}
    \caption{Adjusting hyper-parameters for rgb-SVM and poly-SVM.}
    \label{asjust_hyper_parameters_svm}
\end{figure}

Next, our investigation extends to a more in-depth analysis of the decision boundary. Considering the network has 4 blocks, we experimented with a mix of summation and star operations, applying them in various combinations as demonstrated in Figure~\ref{fig:supp_2dpoints_comprehensive}. The visual results from this exploration suggest that employing star operations in the early blocks of the network yields the most significant benefits.

\noindent\textbf{Impact of hyper-parameters in SVM.} 
It should be noted that parameter adjustments will lead to different visualization results, which may challenge our claim based on Fig.~\ref{fig:2d_points}. However, the change is not significant due to the intrinsic differences between ploy and rbf kernels, as shown
in Fig.~\ref{asjust_hyper_parameters_svm}. Our star-based network decision boundary is consistently similar to the decision boundary of poly-SVM. Hence, not only theoretical proof, our visual experiments also hold firmly.

\section{Exploring Extremely Small Models}
\label{sec:supp_exploring_extremely_small_models}
In this section, we explore the performance of StarNet under extremely small parameters (around 0.5M, 1.0M, and 1.5M). For these extremely small variants, we further tune the MLP expansion ratio besides block number and base embedding width. The detailed configurations of these very small variants are presented in Table~\ref{tab:verysamll_starnet_configurations}.

In this section, we delve into the performance of StarNet when configured with extremely small number of parameters, specifically around 0.5M, 1.0M, and 1.5M. For these ultra-compact variants, we go beyond just adjusting the block number and base embedding width; we also finely tune the MLP expansion ratio. The specific configurations for these very small StarNet variants are detailed in Table~\ref{tab:verysamll_starnet_configurations}.

\begin{table}[]
\centering
\resizebox{\linewidth}{!}{
\begin{tabular}{l|ccc|cc}
\Xhline{3\arrayrulewidth}
Variant    & width & depth  & expand           & Params & FLOPs \\
\Xhline{3\arrayrulewidth}
StarNet-050 & 16 & {[}1, 1, 3, 1{]}  & 3 & 0.54M    & 92.8M   \\
StarNet-100 & 20 & {[}1, 2, 4, 1{]}  & 4 & 1.04M    & 187.1M   \\
StarNet-150 & 24 & {[}1, 2, 4, 2{]}  & 3 & 1.56M    & 229.0M   \\
\Xhline{3\arrayrulewidth}
\end{tabular}
}
\caption{
\textbf{Configurations of very small StarNets}.  We vary the embed width, depth, and MLP expansion ratio to StarNets at 0.5M, 1.0M, and 1.5M parameters.}
\label{tab:verysamll_starnet_configurations}
\end{table}

\begin{table}[]
\centering
    \resizebox{\linewidth}{!}{
    \begin{tabular}{lcccccc}
    \Xhline{3\arrayrulewidth}
    \multirow{2}{*}{\textbf{Model}} & \textbf{Top-1} & {\textbf{Params}} & {\textbf{FLOPs}} & \multicolumn{3}{c}{\bf Latency (ms)} \\
    \cmidrule{5-7}
        &\textbf{(\%)} & \textbf{(M)} & \textbf{(M)} & \textbf{Mobile} & \textbf{GPU} &\textbf{CPU} \\
        \Xhline{3\arrayrulewidth}
        MobileNetv2-050*&66.0	&1.97	&104.5 &0.7 		&1.4	&1.5\\
        \Xhline{3\arrayrulewidth}
        StarNet-050	&\textbf{55.7}	&0.54	&92.8 &\textcolor{black}{\textbf{0.5}} 		&1.0	&1.3\\
        StarNet-100	&\textbf{64.3}	&1.04	&187.1 &\textcolor{black}{\textbf{0.6}} 		&1.3	&2.3\\
        StarNet-150	&\textbf{68.0}	&1.56	&229.0 &\textcolor{black}{\textbf{0.6}} 		&1.5	&2.6\\
    \Xhline{3\arrayrulewidth}
    \end{tabular}
    }
    \caption{\textbf{Performance of extremely small StarNet variants on ImageNet-1k}. All our StarNet are trained with 300 epochs following the training recipe of StarNet-S1. MobileNetv2-050* is taken from the timm library~\cite{rw2019timm}, trained with 450 epochs.}
    \label{tab:compare_imagenet_verysmall}
\end{table} 

The results presented in Table~\ref{tab:compare_imagenet_verysmall} demonstrate that our ultra-compact variants of StarNet are also promising in performance. When compared to MobileNetv2-050, which is trained for longer training period and introduces approximately 25\% more parameters (1.97M \textit{vs.} 1.56M), our StarNet-150 variant still outperforms in both top-1 accuracy and speed on mobile devices.

\begin{figure*}[]
    \centering
    \begin{subfigure}[b]{0.18\linewidth}
        \includegraphics[width=0.9\linewidth]{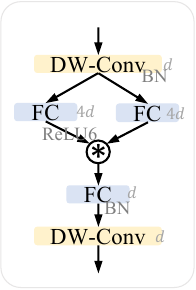}
        \caption{Block I {\scriptsize (default)}}
        \label{fig:supp_block1}
    \end{subfigure}
    \quad
    \begin{subfigure}[b]{0.18\linewidth}
        \includegraphics[width=0.9\linewidth]{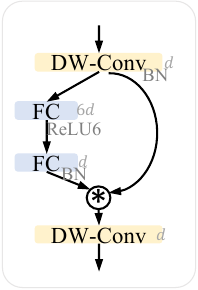}
        \caption{Block II}
        \label{fig:supp_block2}
    \end{subfigure}
    \quad
     \begin{subfigure}[b]{0.18\linewidth}
        \includegraphics[width=0.9\linewidth]{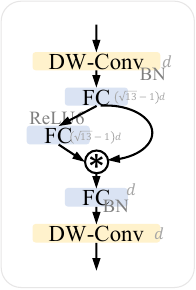}
        \caption{Block III }
        \label{fig:supp_block3}
    \end{subfigure}
    \quad
    \begin{subfigure}[b]{0.18\linewidth}
        \includegraphics[width=0.9\linewidth]{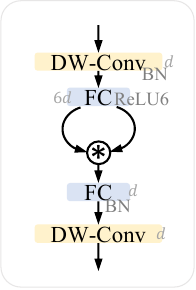}
        \caption{Block IV }
        \label{fig:supp_block4}
    \end{subfigure}
    \quad
    \begin{subfigure}[b]{0.18\linewidth}
        \includegraphics[width=0.9\linewidth]{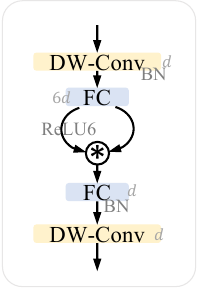}
        \caption{Block V }
        \label{fig:supp_block5}
    \end{subfigure}
\caption{We provide more studies on the block design of StarNet. \textbf{Block I} is the default design used in our StarNet, and the standard star operation as we discussed.  \textbf{Block II} and \textbf{Block III} can be considered as instantiations of \textcolor{orange}{special case II} as presented in Sec.\ref{sec:special_cases}. \textbf{Block IV} and \textbf{Block V} can be considered as instantiations of \textcolor{orange}{special case III}. We vary the expansion ratio to ensure all the block variants have same parameters and FLOPs for fair comparison. Skip connection is ignored for simplicity.}
    \label{fig:supp_block_comprehensive}
\end{figure*}

\section{Activation Analysis}
\label{sec:supp_activation_analysis}
\subsection{Analysis on removing all activations}
In Sec.~\ref{sec:discussions}, we explored the possibility of eliminating all activation functions, presenting preliminary findings in Table~\ref{tab:without_activation} and Table~\ref{tab:activation_placement}. This section delves deeper into the \textbf{detailed analyses of removing all activations}. Utilizing DemoNet as our model for illustration, we provide further results in Table~\ref{tab:demonet_width_supp} and Table~\ref{tab:demonet_depth_supp}.

\begin{table}[]
\resizebox{\linewidth}{!}{
\begin{tabular}{lccccccc}
\Xhline{3\arrayrulewidth}
Width & 96   & 128  & 160  & 192  & 224  & 256  & 288      \\
\Xhline{3\arrayrulewidth}
w/ act. & 57.6 & 64.0 & 68.2 & 71.7 & 73.9 & 75.3 & 76.1  \\
w/o act. & 57.6 & 64.0 & 67.5 & 70.5 & 73.0 & 75.3 & 75.7  \\
 \arrayrulecolor{gray}\hline
 acc. gap  & -  & -  & 0.7$\downarrow$  & 1.2$\downarrow$  & 0.9$\downarrow$  &-  & 0.4$\downarrow$   \\ 
 \Xhline{3\arrayrulewidth}
\end{tabular}
}
\caption{\textbf{Comparison of removing all activations in DemoNet (using star operation) with different widths.} We set the depth to 12. We gradually increase the width by a step of 32. }
\label{tab:demonet_width_supp}
\end{table}

\begin{table}[]
\resizebox{\linewidth}{!}{
\begin{tabular}{lccccccc}
\Xhline{3\arrayrulewidth}
Depth    & 10   & 12   & 14   & 16   & 18   & 20   & 22     \\
\Xhline{3\arrayrulewidth}
w/ act. & 70.3 & 71.8 & 72.9 & 72.9 & 73.9 & 75.4 & 75.4  \\
w/o act. & 69.4 & 70.5 & 72.6 & 73.5 & 74.2 & 74.7 & 75.5  \\
\arrayrulecolor{gray}\hline
acc. gap  & 0.9$\downarrow$  & 1.3$\downarrow$  & \textcolor{orange}{0.3$\uparrow$}  & \textcolor{orange}{0.6$\uparrow$}  & \textcolor{orange}{0.3$\uparrow$}    & 0.7$\downarrow$  & \textcolor{orange}{0.1$\uparrow$}    \\
\Xhline{3\arrayrulewidth}
\end{tabular}
}
\caption{\textbf{Comparison of removing all activations in DemoNet (using star operation) with different depths.} We set the width to 192. We gradually increase the depth by a step of 2.}
\label{tab:demonet_depth_supp}
\end{table}

In most cases, we observe slight performance drop when removing all activations from DemoNet (using star operation). 
However, an intriguing observation emerged: in some instances, the removal of all activations resulted in similar or even better performance. This was notably evident when the $\mathrm{depth}$ values were set to 14, 16, 18, and 22, as detailed in Table~\ref{tab:demonet_depth_supp}. This phenomenon implies that the star operation may inherently provide sufficient non-linearity, akin to what is typically achieved through activation layers. We believe that a more thorough investigation in this direction could yield valuable insights.

\subsection{Exploring activation types}
In our StarNet design, we adopt the ReLU6 activation function, following MobileNetv2. Additionally, we have experimented with various other activation functions, with the results of these explorations presented in Table~\ref{tab:supp_different_acts}. Empirically, we found that StarNet-S4 delivers the best performance when equipped with the ReLU6 activation function.
\begin{table}[]
    \centering
    \resizebox{\linewidth}{!}{
    \begin{tabular}{c@{\extracolsep{4pt}}!{\vrule width 1.0pt}ccccc}
        \multirow{2}{*}{\textbf{Act.}}  & ReLU & GELU & LeakyReLU & HardSwish & ReLU6 \\
         &  {\scriptsize $\mathrm{max}(0,x)$} & {\scriptsize $\frac{x}{2}\left ( 1+\mathrm{erf}\left ( \frac{x}{\sqrt{2}} \right ) \right )$} & {\scriptsize $\mathrm{max}(0.01x,x)$} & {\scriptsize $x\frac{\mathrm{ReLU6}\left ( x+3 \right )}{6}$ }&{\scriptsize $\mathrm{min}(\mathrm{max}(0,x),6)$} \\
         \Xhline{3\arrayrulewidth}
         \textbf{Acc.} & 77.6 & 77.8 & 77.7 & 77.7 & \textbf{78.4} 
        
    \end{tabular}
    }
    \caption{\textbf{Performance of different activations in StarNet-S4.}}
    \label{tab:supp_different_acts}
\end{table}

\section{Exploring Block Designs}
\label{sec:supp_block_analysis}

We provide detailed ablation studies on the design of the StarNet block.  To this end, we present five implementations of star operation in StarNet, as illustrated in Fig.~\ref{fig:supp_block_comprehensive}. Of note is that Block I can be considered as a standard implementation of star operation, Block II and Block III can be considered as instantiations of special case II, which is discussed in Sec.\ref{sec:special_cases}, Block IV and Block V can be considered as different implementations of special case III.  All the block variants have same parameters and FLOPs via varying expansion ratio, ensuring same computational complexity for fair comparison. We test all these 
block variants based on StarNet-S4 architecture, and report the performance in Table~\ref{tab:block_design}. Empirically, we see strong performance for Block I, Block IV, and Block V, while Block II and Block III perform worse. The detailed results suggest that the strong performance stems from the star operation rather than specific block design. We consider the default star operation (Block I) as the essential building block in our StarNet.

We have conducted detailed ablation studies on the design of the StarNet block. To this end, we explored five different implementations of the star operation in StarNet, as depicted in Fig.\ref{fig:supp_block_comprehensive}. Notably, Block I represents a standard implementation of the star operation. Blocks II and III are instantiations of the special case II, discussed in Sec.\ref{sec:special_cases}, while Blocks IV and V offer alternative takes on special case III. All these block variants maintain the same number of parameters and FLOPs by adjusting the expansion ratio, ensuring the same computational complexity for a fair comparison. These block variants were tested within the StarNet-S4 architecture framework, with their performance detailed in Table\ref{tab:block_design}. Empirically, Blocks I, IV, and V demonstrated strong performance, with minimal performance gap. These findings suggest that the effectiveness is more attributable to the star operation itself rather than to the specific design of the blocks. Therefore, we use the default star operation (Block I) as the foundational block in our StarNet.

\begin{table}[]
    \centering
    {
    \small
    \begin{tabular}{ll@{\extracolsep{4pt}}!{\vrule width 1.0pt}c}
        Star Op.&Design &  Top-1\\
        \Xhline{3\arrayrulewidth}
        Standard& Block I    &  78.4\\
        \arrayrulecolor{gray}\hline
        \multirow{2}{*}{Case II}& Block II   &  74.4\\
        & Block III  &  74.4\\
        \arrayrulecolor{gray}\hline
        \multirow{2}{*}{Case III}& Block IV   &  78.5\\
        & Block V    &  \textbf{78.6}\\
    \end{tabular}
    }
    \caption{Performance of different block designs. The detailed implementation of each block can be found in Fig.~\ref{fig:supp_block_comprehensive}.}
    \label{tab:block_design}
\end{table}

\begin{figure}[!h]
    \centering
    \includegraphics[width=1.0\linewidth]{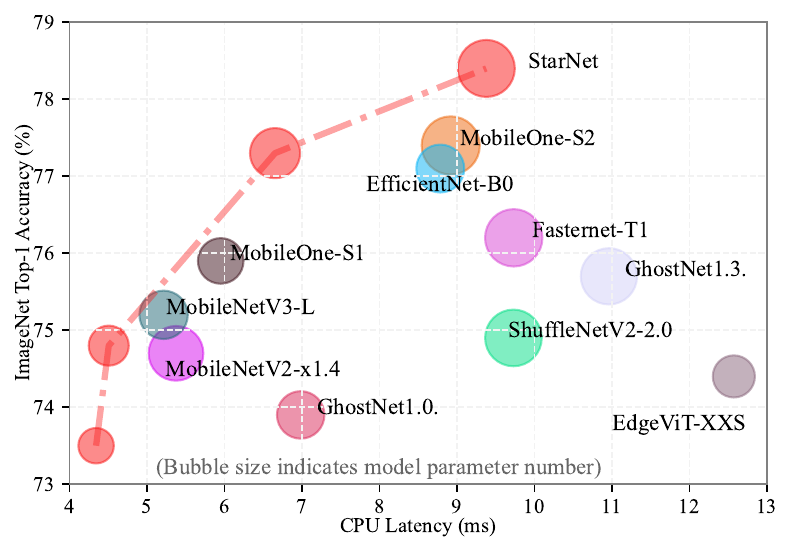}
    \caption{CPU Latency vs. ImageNet Accuracy. Low-accuracy or high-latency models are removed for better visualization.}
    \label{fig:cpu_latency_acc}
\end{figure}
\section{More Latency Analysis on StarNet}
\textbf{CPU latency Visualization.} To better understand the latency of the proof-of-concept model StarNet, we further plot the CPU-latency trade-off in Fig.~\ref{fig:cpu_latency_acc}.

\noindent\textbf{Mobile Device latency Robustness.}  In Table~\ref{tab:compare_imagenet}, we presented the mobile device latency, which is tested on a iPhone13 mobile phone. We further conducted latency testing on 4 different mobile devices, including iPhone12, iPhone12 Pro Max, iPhone 13, and iPhone14,  to test the latency stability of different models. Results in Tab.~\ref{tab:more_iphone_latency} demonstrate that, despite varying latency results for some models, StarNet always shows stable inference across different phones, attributed to its simple design.

\begin{table}[]
\centering
     \caption{\textbf{Supplementary of Table 6}. We further test model latency on four iPhone devices, including iPhone12, iPhone12Pro Max, iPhone13, and iPhone14. The average latency and variance are reported in the last column. We mark most stable models in green color and most unstable models in red.
     }
    \label{tab:more_iphone_latency}
    \vspace{-3mm}
    \resizebox{1.0\linewidth}{!}{
    \begin{tabular}{lcccccr}
        \toprule
    \multirow{2}{*}{\textbf{Model}} & \textbf{Top-1} & \multicolumn{5}{c}{\bf iPhone devices Latency (ms)} \\
    \cmidrule{3-7}
        &\textbf{(\%)} & \textbf{12} & \textbf{12PM} & \textbf{13} & \textbf{14} &\textbf{avg$\pm$std} \\
    
        \toprule
        MobileOne-S0 &71.4  &0.7 &0.7 &0.7 &0.7 &0.7$\pm$0.02 \\
        ShuffleV2-1.0 &69.4 &4.1 &4.1 &0.8 &0.8 &2.4$\pm$\textcolor{red}{1.66} \\
        MobileV3-S0.75 &65.4 &5.5 &5.3 &6.5 &6.6 &6.0$\pm$0.59 \\
        GhostNet0.5 &66.2&10.0 &7.3 &9.7 &8.8 &8.9$\pm$\textcolor{red}{1.05} \\
        MobileV3-S &67.4 &6.5 &5.8 &7.5 &7.7 &6.9$\pm$0.75 \\
        StarNet-S1 &\textbf{73.5} &0.7 &0.7 &0.7 &0.7 &0.7$\pm$\textcolor{green}{0.00} \\
        \midrule%
        MobileV2-1.0 &72.0&0.9 &0.9 &1.0 &0.9 &0.9$\pm$0.04 \\
        ShuffleV21.5 &72.6 &5.9 &5.9 &1.2 &1.2 &3.5$\pm$\textcolor{red}{2.34} \\
        M-Former-52 &68.7 &6.6 &6.3 &8.3 &8.2 &7.4$\pm$0.93 \\
        FasterNet-T0 &71.9 &0.7 &0.7 &0.8 &0.7 &0.7$\pm$0.02 \\
        StarNet-S2 &\textbf{74.8}  &0.7 &0.8 &0.8 &0.8 &0.8$\pm$\textcolor{green}{0.01} \\
        \midrule%
        MobileV3-L0.75 &73.3 &10.9 &11.4 &12.4 &14.2 &12.2$\pm$\textcolor{red}{1.26}  \\
        EdgeViT-XXS &74.4 &1.8 &1.8 &1.2 &1.2 &1.5$\pm$0.31 \\
        MobileOne-S1 &75.9 &0.9 &0.9 &0.9 &0.9 &0.9$\pm$0.03  \\
        GhostNet1.0&73.9 &7.9 &7.4 &10.1 &10.2 &8.9$\pm$\textcolor{red}{1.28}  \\
        EfficientNet-B0 &77.1 &1.6 &1.5 &1.7 &1.7 &1.6$\pm$0.10 \\
        MobileV3-L &75.2 &11.4 &12.9 &15.1 &14.9 &13.6$\pm$\textcolor{red}{1.52} \\
        StarNet-S3 &\textbf{77.3} &0.9 &0.9 &0.9 &0.9 &0.9$\pm$\textcolor{green}{0.02} \\
        \midrule%
        EdgeViT-XS &77.5 &3.5 &3.5 &1.6 &1.6 &2.5$\pm$0.96 \\
        MobileV2-1.4&74.7 &1.1 &1.2 &1.3 &1.3 &1.2$\pm$0.08 \\
        GhostNet1.3 &75.7 &9.7 &9.0 &12.4 &12.5 &10.9$\pm$\textcolor{red}{1.57} \\
        ShuffleV2-2.0 &74.9 &19.9 &16.9 &1.8 &1.5 &10.0$\pm$\textcolor{red}{8.44} \\
        Fasternet-T1 &76.2 &0.9 &0.9 &1.0 &1.0 &1.0$\pm$\textcolor{green}{0.03} \\
        MobileOne-S2 &77.4 &1.0 &1.1 &1.2 &1.2 &1.1$\pm$0.07 \\
        StarNet-S4 &\textbf{78.4} &1.0 &1.0 &1.1 &1.1 &1.1$\pm$\textcolor{green}{0.03} \\
\bottomrule
    \end{tabular}
    }
   \vspace{-5mm}
\end{table}

\end{document}

%% file: preamble.tex
\usepackage[dvipsnames]{xcolor}